\documentclass{article}
\usepackage{arxiv}

\usepackage{graphicx} 
\usepackage{todonotes}

\usepackage[noend]{algpseudocode}
\usepackage{verbatim} 
\usepackage{marginnote} 
\usepackage{longtable} 
\usepackage{booktabs}
\usepackage{amsmath}
\usepackage[hidelinks]{hyperref}
\usepackage{threeparttable}
\usepackage{xcolor}
\usepackage{float}
\usepackage[normalem]{ulem}
\usepackage{makecell}

\usepackage{xcolor}
\usepackage{amsmath}
\usepackage{algorithm}

\usepackage{adjustbox}

\NewDocumentCommand{\orcid}{m}{%
  \href{https://orcid.org/#1}{%
    \includegraphics[height=1.6ex]{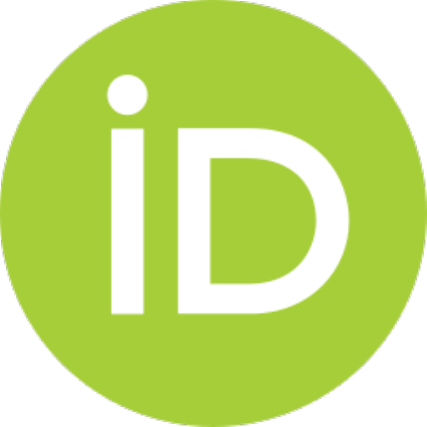}%
  }%
}

\NewDocumentCommand{\mail}{m}{%
  \href{mailto:#1}{\texttt{#1}}%
}

\title{DCV-ROOD Evaluation Framework: Dual Cross-Validation
for Robust Out-of-Distribution Detection}

\author{\orcid{0009-0005-4724-0671}
    Arantxa Urrea-Castaño$^{*,a,b}$ \\
    \mail{s.arantxa.uc@ugr.es}
    \And
    \orcid{0009-0002-1413-2277}
    Nicolás Segura-Kunsagi$^{a,b}$ \\
    \mail{nkunsagi@ugr.es}
    \And
    \orcid{0000-0001-8587-4345}
    Juan Luis Suárez-Díaz$^{a,b}$ \\
    \mail{jlsuarezdiaz@decsai.ugr.es}
    \And
    \orcid{0000-0002-0183-044X}
    Rosana Montes$^{a,c}$ \\
    \mail{rosana@ugr.es}
    \And
    \orcid{0000-0002-7283-312X}
    Francisco Herrera$^{a,b}$ \\
    \mail{herrera@decsai.ugr.es}%
}

\newcommand{\affiliations}{
  \vspace{-3em}
  \begin{center}
    \textsuperscript{a} \href{https://dasci.es}{Andalusian Institute of Data Science and Computational Intelligence (DaSCI), University of Granada, Granada, Spain} \\
    \textsuperscript{b} \href{https://decsai.ugr.es}{Department of Computer Science and Artificial Intelligence (DECSAI), University of Granada, Spain} \\
    \textsuperscript{c} \href{https://lsi.ugr.es}{Department of Software Engineering (LSI), University of Granada, Spain} \\
    \textsuperscript{*} {Corresponding Author}
    \vspace{3.5em}
  \end{center}
}

\begin{document}

\maketitle
\affiliations

\begin{abstract}
Out-of-distribution (OOD) detection plays a key role in enhancing the robustness of artificial intelligence systems by identifying inputs that differ significantly from the training distribution, thereby preventing unreliable predictions and enabling appropriate fallback mechanisms. Developing reliable OOD detection methods is a significant challenge, and rigorous evaluation of these techniques is essential for ensuring their effectiveness, as it allows researchers to assess their performance under diverse conditions and to identify potential limitations or failure modes. Cross-validation (CV) has proven to be a highly effective tool for providing a reasonable estimate of the performance of a learning algorithm. Although OOD scenarios exhibit particular characteristics, an appropriate adaptation of CV can lead to a suitable evaluation framework for this setting. This work proposes a dual CV framework for robust evaluation of OOD detection models, aimed at improving the reliability of their assessment. The proposed evaluation framework aims to effectively integrate in-distribution (ID) and OOD data while accounting for their differing characteristics. To achieve this, ID data are partitioned using a conventional approach, whereas OOD data are divided by grouping samples based on their classes. Furthermore, we analyze the context of data with class hierarchy to propose a data splitting that considers the entire class hierarchy to obtain fair ID-OOD partitions to apply the proposed evaluation framework. This framework is called Dual Cross-Validation for Robust Out-of-Distribution Detection (DCV-ROOD). To test the validity of the evaluation framework, we selected a set of state-of-the-art OOD detection methods, both with and without outlier exposure. The results show that the method achieves very fast convergence to the true performance.

\end{abstract}


\newpage
\section{Introduction}


Artificial Intelligence (AI) is increasingly present in our lives, acting as a transformative technology in the world~\cite{herrera2025responsible}. The ability to streamline and automate processes has an impact in different areas. Due to its widespread use in different disciplines, it is necessary to ensure that AI systems are safe, ethical and positive for society~\cite{hendrycks2025ai}.

As the deployment of machine learning systems in real-world applications becomes increasingly widespread, concerns around AI safety and reliability have gained prominence. 
Creating a responsible AI system requires, among other things, robustness to unexpected events, often referred to as black swans~\cite{taleb2007blackswans}. Ensuring that models behave as intended, particularly in the presence of distributional shifts, rare scenarios, or imperfect data, is essential for the safe integration of AI into high-stakes domains~\cite{hendrycks2021unsolved}.

In this context, one of the main challenges faced by modern AI is the detection of out-of-distribution (OOD) data~\cite{yang2021generalized}. This issue is part of broader efforts to build safer and more robust systems, aiming to identify data that does not belong to the distribution on which the AI model was trained. The goal is to filter out such data before it is processed by the model, preventing potentially inappropriate responses. OOD detection plays a fundamental role in fields like autonomous driving~\cite{henriksson2023evaluation}, medical diagnosis~\cite{tamo2023uncertainty,wei2025fine}, natural language processing~\cite{zheng2020out}, and image/video recognition~\cite{liu2025iw,liu2020energy}, among others. Without effective detection, erroneous, unknown, or malicious inputs can lead to inaccurate or faulty model outputs, which could have serious consequences.

The process of building an OOD detector for an AI model consists of two main phases, as in typical learning models: a training stage and a prediction stage. In the training stage, the model is provided with all the training data, known as the in-distribution (ID) set. During the prediction stage, the detector receives any type of data and must determine whether they are ID or OOD. Some detectors are designed to train with a subset of OOD data (denoted as outlier exposure)~\cite{hendrycks2018deep}, assuming that the OOD data encountered during the prediction phase may differ significantly from the training OOD data. Other detectors, by contrast, are developed without relying on any OOD data during training, and instead operate solely based on the ID set. Both approaches have been extensively studied in the literature. There are various approaches to develop OOD detectors~\cite{yang2021generalized}, including classification-based, distance-based, density-based, reconstruction-based, and hybrid methods, among others.

Beyond the design of OOD detection techniques, an equally important aspect is the rigorous evaluation of detector robustness. This evaluation requires extensive experimentation across multiple datasets. Some limitations in OOD experimentation arise from the need to acquire large volumes of labeled data~\cite{zhang2025unsupervised} or from the potential semantic similarity between marginal OOD samples and ID data~\cite{long2024rethinking}. In this paper, we focus on the necessity of robust validation of detectors to reduce variability and reliably approximate the true OOD detection error. Testing detectors on different datasets alone is not sufficient, as their performance can vary significantly even within the same dataset, depending on how the training and testing splits are constructed. To obtain more reliable and generalizable results, experiments should incorporate varied sampling designs, allowing for fairer and more comprehensive assessments. Nevertheless, conducting experiments that combine multiple datasets and sampling strategies entails a high computational cost. As a result, computational efficiency is often prioritized, even at the expense of compromising the rigor of the evaluation.

Cross-validation (CV) has proven to be an efficient and robust method to asses model performance and generalization ability~\cite{yadav2016analysis}. Through CV techniques, models are evaluated on different samples from the same dataset, which enables greater reliability in the results without requiring a large number of experiments or incurring high computational costs. However, to date, no specific scheme has been proposed for its application in the context of OOD detection. The conventional application of CV techniques is not suitable in this setting, as experimentation requires datasets that include both ID and OOD samples. When applied in the conventional way, these techniques expose the model to OOD classes during training, effectively turning them into ID and undermining the validity of OOD detector evaluation. By adapting a CV design that adequately addresses both ID and OOD data, it is possible to extend the applicability of CV to OOD detection.

This paper proposes a new evaluation framework designed specifically for OOD detection problems. This framework consists in a variation of CV, combining two approaches to handle ID and OOD data, respectively: a stratified CV for ID classes and a group CV for OOD classes. In each iteration, the detector is trained using an ID set that maintains the original class proportions, along with an OOD set that varies in each iteration and is always different during testing. Given the dual nature of the partitions generated by the algorithm for each dataset, we have named this evaluation framework DCV-ROOD \emph{(Dual Cross-Validation for Robust OOD Detection}). Furthermore, we propose a DCV-ROOD framework specifically designed for scenarios where the problem involves a hierarchy of classes or superclasses, aiming to create representative partitions aligned with the objectives of the problem. While CV has been explored in the development of methods such as ensembles for OOD detection~\cite{liu2023ood}, this is the first evaluation framework specifically proposed for OOD problems, to the best of our knowledge. 

To assess the effectiveness of DCV-ROOD, a reference \emph{benchmark truth} was built by applying various OOD detection methods across multiple scenarios and identifying statistically significant differences between them. DCV-ROOD framework is then applied to evaluate whether it can replicate those same distinctions. Agreement scores reached up to 9.8571, where a score of 10 indicates that the DCV-ROOD detected exactly the same significant differences as the benchmark truth.

The results indicate that the DCV-ROOD evaluation framework has the potential to enhance the robustness of OOD detection method evaluations.

This work is structured as follows: Section~\ref{sec:background} provides the general context associated with OOD detection and CV. Section~\ref{sec:proposal} describes the DCV-ROOD framework in detail. Sections~\ref{sec:experiments} and~\ref{sec:analysis} introduce the experiments conducted to demonstrate the statistical validity of DCV-ROOD and analyzes the results obtained. Finally, Section~\ref{sec:conclusions} presents the conclusions and highlights the contributions and final insights of this work.

\section{Background} \label{sec:background}

In this section, the importance of understanding both OOD detection techniques and CV methods is emphasized, as each plays a significant role in improving the reliability and robustness of machine learning models. OOD detection enables models to recognize and respond appropriately to unfamiliar inputs, while CV provides a foundation for more accurate and unbiased model evaluation. Given their independent yet complementary contributions, this section is structured in two parts: the first offers an overview of generalized OOD detection, examining key approaches based on their goals, implementation strategies, and underlying methodologies; the second outlines the fundamentals of CV, including the challenges it addresses and the main solutions that have been developed around it.

\subsection{Out-of-distribution} \label{back_ood}
Image classification models aim to determine the category of a new image. When the model has been trained under the closed-world hypothesis, it is assumed that the model knows all possible situations that it will face. Whereas when the open world hypothesis is assumed, the model knows that it is possible that it will face unknown situations (for example, new classes, changes on the clases, etc.)~\cite{zhou2022open}. Assuming a closed-world hypothesis poses the risk that the model may incorrectly classify a novel instance by assigning it to one of the known classes, despite the instance not belonging to any of them. Through the open-world hypothesis an image classification model should be able to distinguish whether an image is within the distribution of known classes ID, or not OOD~\cite{yang2021generalized}.

Generalized OOD detection encompasses several related tasks, including anomaly detection (AD), novelty detection (ND), outlier detection (OD), open-set recognition (OSR), and OOD detection itself~\cite{yang2021generalized}. While these tasks share overlapping goals, primarily identifying data that differ from the training distribution, they differ in methodology and evaluation. For instance, OSR aims to detect semantic anomalies while classifying ID data, but it discourages the use of external data to maintain its core assumption of open space risk minimization~\cite{boult2019learning}. In contrast, OOD detection supports using entirely different datasets to enhance robustness and coverage, and can incorporate broader learning paradigms like multi-label classification and density estimation~\cite{salehi2022unified,yang2021generalized}. Specifically, this research focuses on semantic anomalies, that is, images that do not belong to the model's training classes and constitute a type of OOD data.

Within semantic anomalies, according to the degree of semantic similarity between the OOD images and the known classes~\cite{ren2021simple}, two types of OOD detection tasks are distinguished: near-OOD and far-OOD. Near-OOD tasks are more challenging because they involve classes that are semantically more similar to the ID classes, whereas far-OOD tasks are simpler due to their greater dissimilarity from the ID classes~\cite{winkens2020contrastive}.

Recent advances, particularly in image classification, have led to the development of effective techniques that demonstrate their efficacy across diverse datasets. This progress also reflects the growing importance of OOD detection in research aimed at developing reliable and safe AI systems.  Several strategies have been proposed for OOD detection, and they can be grouped into distinct categories as outlined in the literature~\cite{yang2021generalized}. The following list provide an overview of these approaches.

\begin{itemize}
    \item Density-based methods: these methods model the distribution of ID data and identify OOD samples as those falling in low-density regions~\cite{hendrycks2022scaling}. They include parametric approaches (e.g., multivariate Gaussian models) and non-parametric ones (e.g., kernel density estimation). Some techniques also use generative models or energy-based scores to detect low-probability inputs~\cite{liu2020energy}.
     
    \item Classification-based methods: these approaches rely on the outputs of classification models, such as softmax probabilities, to estimate whether a sample belongs to the known ID classes. They are simple to implement and do not require retraining the model~\cite{hendrycks2017baseline,liu2023gen,park2023nearest,sun2022out,miao2025opencil}.

    \item Distance-based methods: these approaches assess how far a sample is from the training class centers or prototypes. They employ distance metrics such as Mahalanobis, Euclidean, or nearest neighbors, as well as cosine similarity or distances in latent feature spaces, to distinguish OOD instances~\cite{NEURIPS2018_abdeb6f5,liu2023fast,sevillano2025stood,jia2025enhancing}.
   
\end{itemize}
In addition to the approaches described above, two other relevant categories include reconstruction-based methods~\cite{yang2021generalized} and theoretical analyses~\cite{zhang2021understanding}. The former rely on the ability of models to accurately reconstruct ID data, using reconstruction errors to detect anomalies. The latter focus on providing formal guarantees and theoretical insights into the behavior and limitations of OOD detection methods.

In Section~\ref{sec:experiments}, several of these techniques will be employed to asses the effectiveness of the DCV-ROOD evaluation framework. 

\subsection{Cross-validation}
The assessment of the quality of a model, or that of a set of learning models, is of great importance, as it provides a measure of performance. Without proper estimation, it is impossible to predict how the model will behave once deployed. A common practice for obtaining a reasonable estimate is to sample a different dataset from the one used for training and evaluate the model on this new set. This dataset is typically referred to as the validation or test set.

It can be theoretically shown, using Hoeffding's inequality~\cite{shalev2014understanding}, that for any $\delta \in ]0,1[$, with a probability of at least $1-\delta$, the difference between the estimation on the test set, $L_{\text{T}}(h)$, and the true distribution, $L_D(h)$, for a given learner $h$, can be bounded as

\[ |L_T(h) - L_D(h)| \le \sqrt{\frac{\text{log}(2/\delta)}{2|T|}}, \]

where $T$ denotes the test set and $|T|$ its cardinality. For a fixed $\delta$, it can be observed that this bound depends solely on $|T|$. Consequently, as long as a sufficiently large test set can be sampled, the true performance of the learner can be estimated with arbitrary precision, and with probability $1-\delta$. 

However, in many practical scenarios, data is quite limited, and it is not feasible to reserve large amounts of data for validation purposes. In this context, CV is proposed as an alternative technique that maximizes the use of available data, reduces computational cost, and enables fairer model evaluation. It involves the creation of multiple partitions of the data, which alternate between training and testing in several iterations.

Various validation techniques have been developed to assess model performance and generalization throughout the history of machine learning \cite{yadav2016analysis}. From early approaches such as resubstitution and holdout validation to more sophisticated methods, this evolution reflects an ongoing effort to improve the reliability of model evaluation.

Resubstitution involves assessing a model on the same data used for training. While historically important as one of the simplest validation strategies, it typically yields overly optimistic performance estimates. The introduction of holdout validation addressed this limitation by splitting data into separate training and test sets. Despite its simplicity, holdout validation remains sensitive to the specific partitioning of data.

To improve robustness, CV was introduced as a systematic extension of these ideas. Stone \cite{stone1974cross} and Lachenbruch \cite{lachenbruch1968estimation} refined prediction assessment through CV, building on earlier ideas by Mosteller and Wallace (1963) \cite{mosteller1963inference}. Since the 1960s, with contributions from Mosteller and Tukey \cite{mosteller1968data} and many others, CV has undergone numerous refinements and has become a fundamental component of model validation in machine learning.

In $k$-fold CV, the dataset is divided into $k$ partitions; in each of $k$ iterations, one partition serves as the test set while the remaining partitions are used for training. The final error estimate is computed as the average of the prediction errors across all folds. A special case of this approach is leave-one-out CV, where $k$ equals the number of data points, training on all data except one instance per iteration. Although exhaustive, this variant is computationally expensive and often exhibits higher variance.

Additional strategies such as repeated holdout, repeated CV, and Monte Carlo CV \cite{xu2001monte} increase robustness by repeating data splits multiple times. Bootstrapping \cite{kohavi1995study} estimates performance by sampling with replacement to generate multiple training and testing sets. Nested CV \cite{cawley2010over} has become essential for hyperparameter tuning, helping to mitigate data leakage. Stratified CV \cite{zeng2000distribution} maintains class balance in folds, which is especially important for imbalanced datasets. Distribution-balanced and distribution-optimally-balanced stratified CV (DB-SCV and DOB-SCV) \cite{moreno2012study} aim to minimize covariate shift by ensuring folds remain as similar as possible. Group CV \cite{pedregosa2011scikit} preserves group structure by keeping related observations together in train/test splits to prevent leakage. Time-series CV \cite{bergmeir2012use} enforces temporal ordering to respect the sequential nature of the data. The efficiency of validation methods is also a recurrent concern, with the goal of exploiting properties of specific learning algorithms to develop fast validation procedures in those scenarios~\cite{an2007fast,lau2004leave}.

Today, CV and its many variants remain the standard for robust model evaluation across diverse domains, with adaptations designed to address the specific characteristics and challenges of each problem.

When estimating the generalization error in CV, both the bias and the variance of the estimates must be considered~\cite{hastie2009elements}. As $k$ increases, the bias decreases because the training partitions become more similar to the full dataset. In the case of leave-one-out CV, this leads to approximately unbiased estimates. However, as $k$ increases, the variance of the estimates also increases (unless a sufficiently large dataset is available~\cite{cawley2003efficient}), 
as the models are also being trained on very similar subsets of the data~   \cite{kohavi1995study,jiang2017error}. Thus, it is crucial to strike a balance between bias and variance in order to obtain reliable estimates. Another important consideration is the computational cost, which grows with $k$. Empirical analyses typically suggest that values of $k=5$ $k=10$ are reasonable, balancing bias, variance, and computational expense~\cite{rodriguez2009sensitivity}. Under certain assumptions of locality or stability of the estimators, it is also possible to bound the difference of the errors between the true estimation and the CV estimation~\cite{ kearns1997algorithmic,rogers1978finite}.

CV has been widely employed in classification tasks. Within this context, two notable variants are worth highlighting:

\begin{itemize}

    \item \emph{Stratified $k$-fold CV:} this variant is particularly useful in classification problems with significant class imbalance. The sampling strategy ensures that each fold preserves the original class distribution, maintaining similar proportions of each class across all folds. This leads to more reliable performance estimates, especially for minority classes. Even in balanced datasets, stratification helps maintain consistent class proportions across folds, contributing to a fairer evaluation.

    \item \emph{Group $k$-fold CV:} this variant is appropriate when the data contains predefined groups or categories. It ensures that all instances belonging to the same group are kept within the same fold. This prevents the model from inadvertently learning from specific groups that should be excluded during certain training-test splits, thereby allowing for more controlled and interpretable evaluation depending on the grouping structure.

\end{itemize}

Although many CV techniques exist, they are not always immediately applicable to new problems in different machine learning tasks. As research in machine learning progresses, the challenges for techniques aimed at improving robustness also increase. CV is one such example, which once represented a advancement in classification tasks. In particular, applying the DCV-ROOD is not an immediate procedure. CV techniques must be properly adapted to each segment of the dataset, and existing techniques need to be developed.

\section{Dual Cross-Validation for Robust Out-of-Distribution Detection (DCV-ROOD)}\label{sec:proposal}
\label{c_v_ood}

CV offers a cost-effective alternative for supervised learning, yet has not been adapted for OOD detection. Conventional CV risks exposing the model to OOD data during training, potentially including classes that will also be used in the test set, thereby undermining evaluation integrity.

This section introduces the novel evaluation framework DCV-ROOD, which implements a dual CV scheme. The term \emph{dual} refers to the use of complementary CV strategies for ID and OOD data. This differentiation allows ID and OOD data to be handled separately, preserving class separation to prevent contamination between training and testing sets, a prerequisite for effective outlier exposure. Such isolation is essential to ensure a reliable and robust evaluation of OOD detection models.

\subsection{DCV-ROOD for non-hierarchical classes} \label{ssec:cv_ood}
We refer to a dataset with non-hierarchical classes as one in which no inherent hierarchical structure exists, or such structure is not taken into account when designing the folds. In order to perform DCV-ROOD, ID and OOD classes are assigned to their respective folds, which are then appropriately combined for training and testing purposes.

\begin{itemize}
    \item For ID classes, there are two main options for sampling using $k$-fold CV:
    \begin{itemize}
      \item Standard $k$-fold (simple random sampling): this could be a suitable choice for balanced datasets (i.e., when all classes have equal sample sizes). By applying simple random sampling, each class would retain the same probability of selection, allowing its representation in each fold to closely match that of the original dataset.

     \item Stratified $k$-fold (stratified sampling): this alternative may be more appropriate when dealing with imbalanced datasets, as it would approximately preserve the class distribution within each fold, maintaining proportional representation.
    \end{itemize}

    \item For OOD classes, one main method can be considered to ensure a proper evaluation setup: group $k$-fold. In settings where outlier exposure is employed, it is important to ensure that the model does not learn the OOD classes in advance. Otherwise, those classes would effectively become ID classes, which would defeat the purpose of the OOD evaluation. To mitigate this, group $k$-fold could be explored as a method, as it ensures that instances from the same class are not split across different folds, thereby preventing data leakage between training and testing. 
    
    In the absence of outlier exposure, standard $k$-fold or stratified $k$-fold can be used. In this case, regardless of the OOD fold selected for testing, there is no risk of the model learning those classes, as they are never introduced during training. The choice between standard and stratified $k$-fold depends on class balance. In any case, within a general experimental setting where both methods with outlier exposure and those without it must be compared, the group k-fold partitioning is the most inclusive method, as it enables a fair comparison of all methods on the same data split.

\end{itemize}

\begin{figure}[H]
    \makebox[\textwidth][c]{%
        \includegraphics[width=1\textwidth]
        {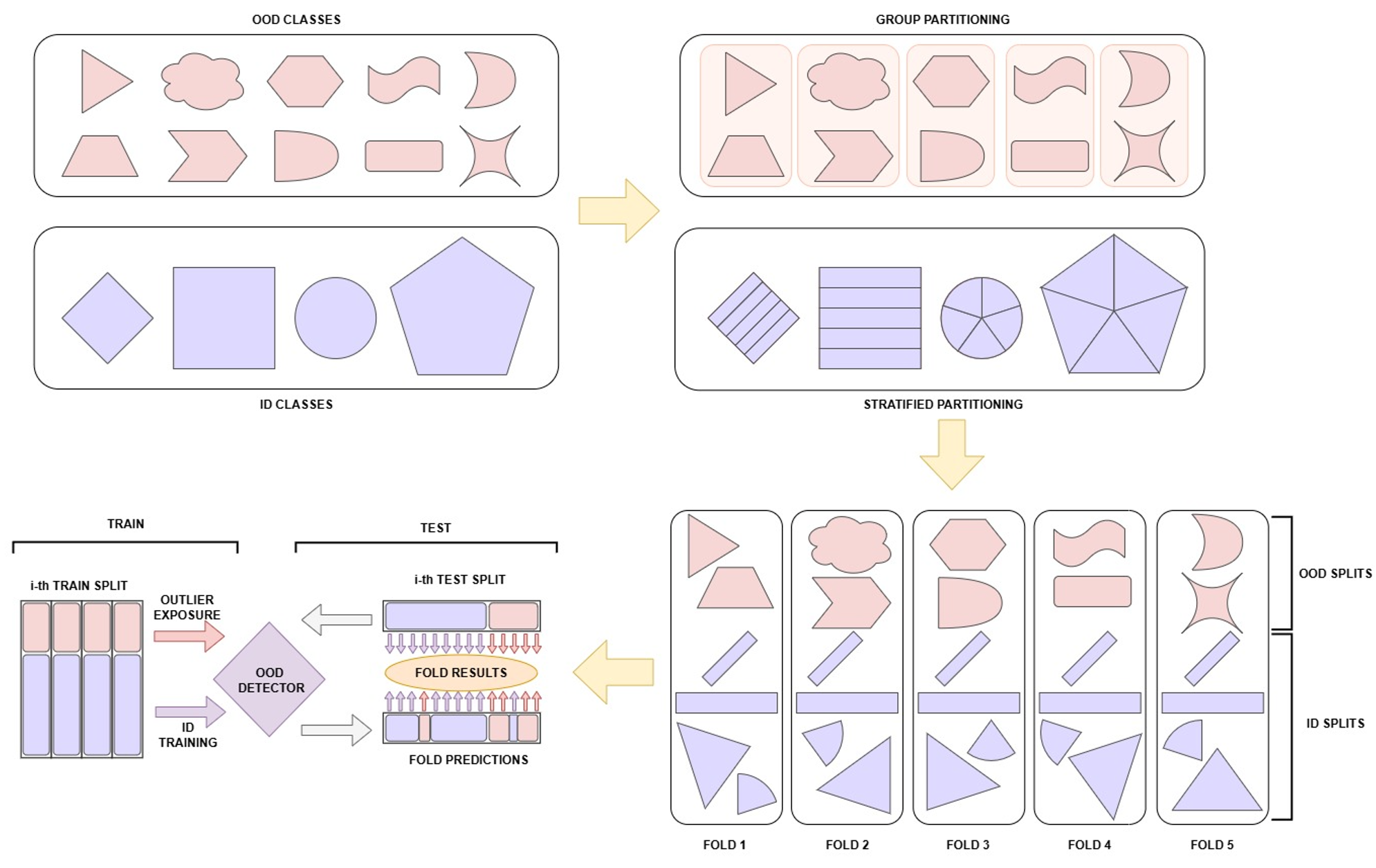}%
    }
    \caption{Illustration of DCV-ROOD used in an outlier exposure scenario.}
    \label{fig1}
\end{figure}

Figure~\ref{fig1} illustrates an example in which training is conducted under an OOD-aware setting. A 5-fold DCV-ROOD is applied to an OOD detection task, aiming to perform outlier exposure during training. The evaluation framework can then be divided into the steps shown in Figure~\ref{fig1}:
\begin{enumerate}
    \item We start from the full dataset, containing ID and OOD classes, on which we aim to evaluate the OOD detection algorithms. 
    
    \item ID data are separated using stratified sampling, while OOD classes are divided into distinct groups.
    \item CV folds are generated. In each fold, one group of OOD classes is included along with a balanced stratified subset of each ID class obtained in the previous step.
    \item The detector is trained and tested. When training with outlier exposure, the model is fed with the ID and OOD data from the four training folds. For prediction, the held-out test fold containing both ID and OOD data is used. Detection results are then compared with the ground truth to compute the evaluation metrics.
\end{enumerate}

Building on the DCV-ROOD evaluation framework presented in this section, the next part focuses on its application to hierarchical classes.

\subsection{DCV-ROOD for hierarchical classes}
\label{hierarchical}
In many real-world classification problems, classes exhibit a hierarchical structure. In such contexts, DCV-ROOD must be adapted to the hierarchy of the dataset to avoid yielding unreliable estimates. If applied directly as described in subsection~\ref{ssec:cv_ood}, without considering the hierarchical structure when defining the ID and OOD folds, the results may be misleading.

This issue is particularly relevant in domains such as healthcare, where a disease may be divided into subtypes (as in the MIMIC-III dataset~\cite{johnson2016mimic}), or in computer vision, where broad categories like ``birds” comprise multiple species (as in iNaturalist~\cite{vanhorn2018inaturalist} or CIFAR-100~\cite{netzer2011reading}, which organize classes across multiple levels). In these cases, DCV-ROOD, by adapting to the hierarchy of the dataset, enables a more rigorous evaluation of OOD detection models.

In the case of a dataset with class hierarchy, there exists an implicit or explicit structure that groups data into two or more levels reflecting relationships among them (for example, superclass → class → subclass). The hierarchical structure of the dataset introduces risks in the creation of ID and OOD splits during experimentation, and the same applies when performing DCV-ROOD if the hierarchy of the dataset is not considered. There are two critical stages. The first is the selection of classes to be assigned to the ID and OOD splits, when this assignment is not predetermined. The second is the construction of the ID and OOD folds for their respective splits. Inadequate sampling during the creation of the folds can prevent the model from fully capturing the variability of each class, increasing the likelihood of misclassifying valid ID examples as OOD (false OOD), thereby reducing the reliability of OOD detection. Similarly, if the OOD test folds do not accurately reflect the hierarchy of the dataset, the detection task can become artificially easy, producing evaluations that fail to capture the true semantic complexity of the data.

By means of DCV-ROOD, the hierarchical levels are considered during the construction of the ID and OOD folds, enabling both the training and testing processes to more accurately capture the underlying data structure, thereby yielding more reliable and generalizable results.

\subsubsection*{Illustrative example of a dataset with class hierarchy}
Let us consider the following example with three hierarchical levels. The example presents a hierarchical animal dataset organized according to the following class hierarchy: 

\begin{itemize}
\item Level 1 (superclass): \textcolor{violet}{Mammals}, \textcolor{teal}{Birds}, etc.
\item Level 2 (class): \textcolor{violet}{Canines, Felines, etc.} / \textcolor{teal}{Aquatic Birds, Raptors, etc.}
\item Level 3 (subclass): \textcolor{violet}{Cocker, Labrador, Shih Tzu, Bombay Cat, etc.} / \textcolor{teal}{Penguins, Swan, Owl, etc.}
\end{itemize}

\begin{figure}[H]
    \centering
    \includegraphics[width=0.7\textwidth]{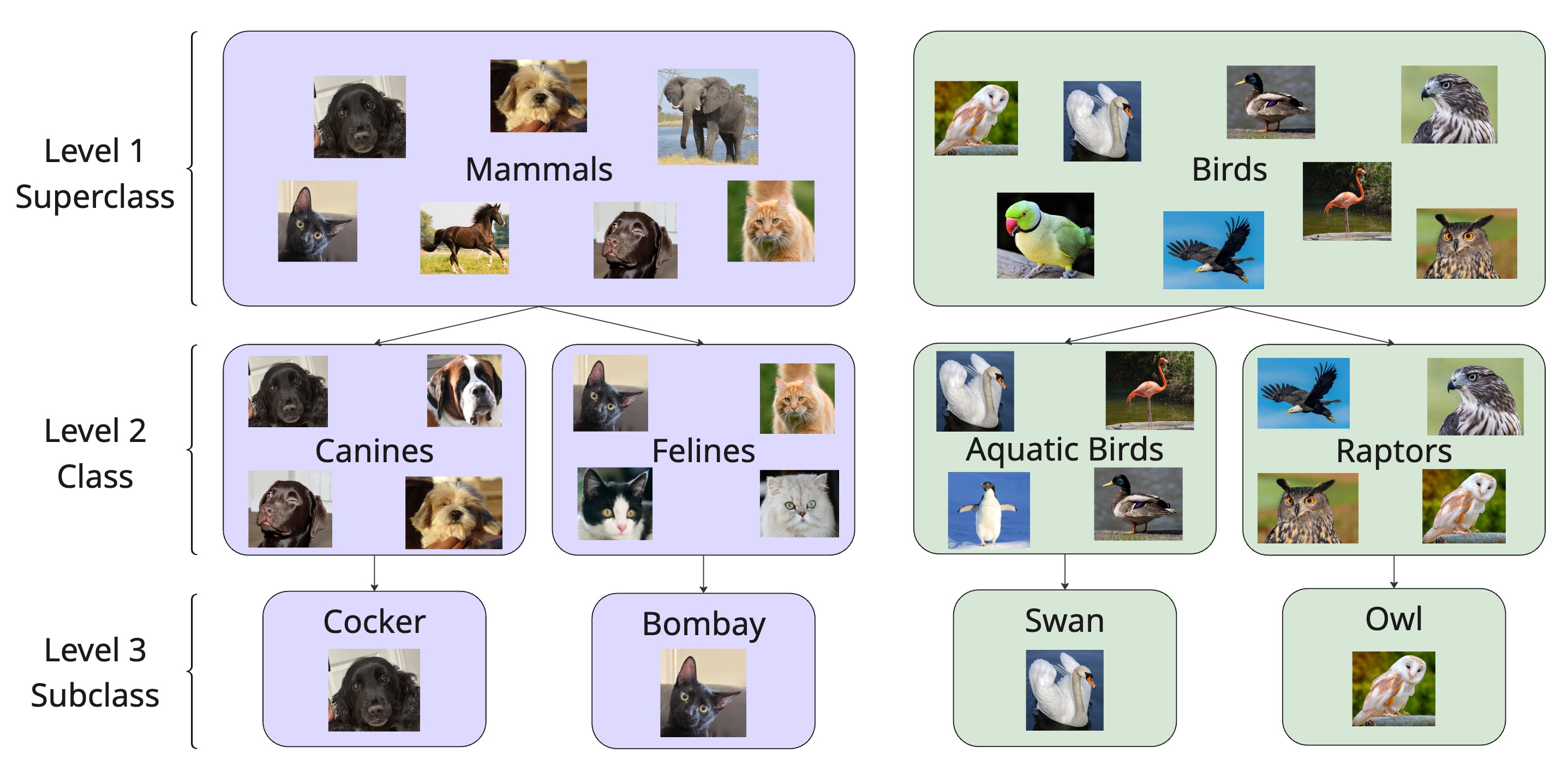}
    \caption{A simplified view of the category hierarchy is shown, featuring two superclasses, four classes, and four subclasses for illustration.}
    \label{fig:nonhierarcquical partition}
\end{figure}

In the hierarchy above, colors are used solely to illustrate the hierarchical relationships (e.g., \textcolor{violet}{violet} for mammals and their subcategories, \textcolor{teal}{teal} for birds). A visual representation of this structure is provided by way of example in Figure~\ref{fig:nonhierarcquical partition}.

If, during training (ID), the model sees some sub-subtypes of \textcolor{violet}{\textit{Canines}} such as \textcolor{violet}{\textit{Cocker}} and \textcolor{violet}{\textit{Labrador}}, and in the test set only \textcolor{teal}{\textit{Birds}} are used as OOD, the task becomes relatively easy. ``Birds'' are a superclass with high semantic difference, and detecting them as OOD is more trivial.

However, if we include \textcolor{violet}{\textit{Husky}} in the OOD test (which also belongs to \textcolor{violet}{\textit{Canines}}), the problem becomes significantly more challenging. In this case, the model is being tested within the same superclass and class it saw during training, with variation only at the most specific level. This setup truly evaluates the model’s ability to detect subtle differences between highly similar instances. 

When considering only the differences between higher-level classes in the hierarchy, there is a risk of overestimating the model’s ability to detect OOD data. The real challenge lies in detecting differences across all class levels, that is, among subclasses within the same class. Thus, it is crucial that the ID/OOD split considers this granularity in order not to overestimate the model's performance and to simulate more realistic scenarios.

\begin{figure}[H]
    \centering
    \includegraphics[width=0.8\textwidth]{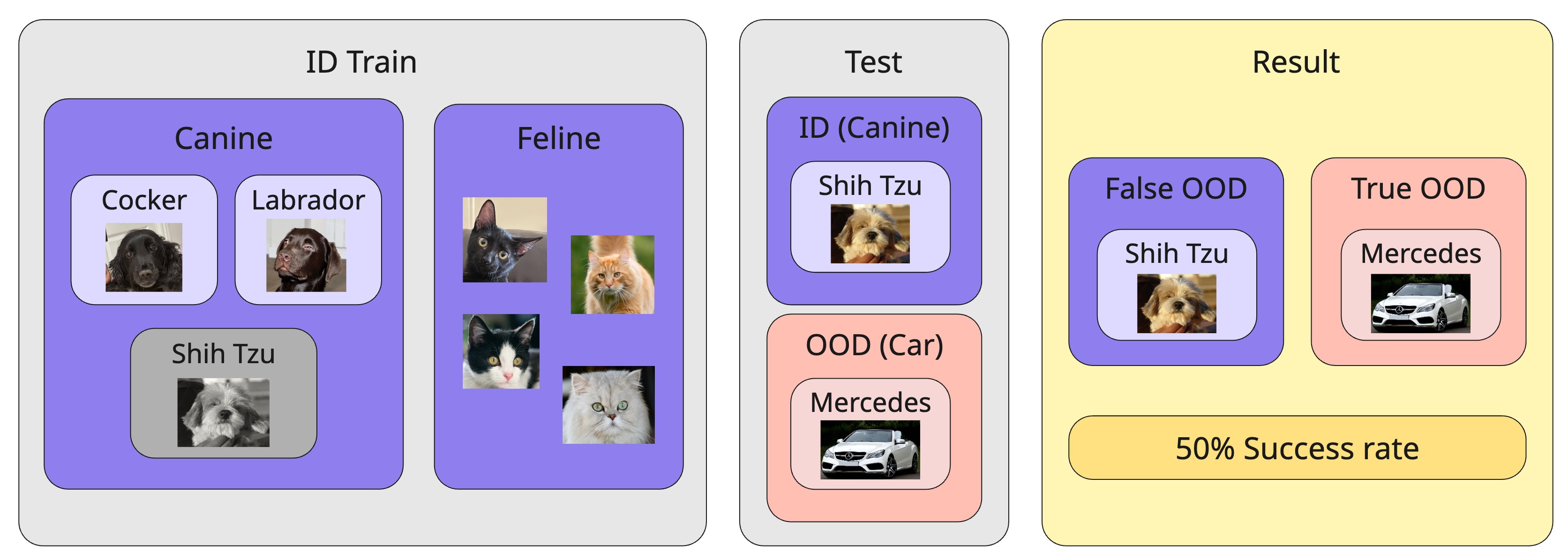}
    \caption{Visualization of a False OOD due to missing subclass representation in training. Although the model learns at the class level, unseen subclasses like Shih Tzu (shown in gray) may be flagged as OOD at test time, even if they belong to an ID class.
    }
    \label{ejem_error}
\end{figure}

In hierarchical classes, even when we do not explicitly intend to classify the lower levels of the hierarchy (such as subclasses in the example), their existence must be considered when preparing folds for DCV-ROOD. Continuing with the example, suppose we are working with a dataset including mammals and birds and aim to train an OOD detection model that only distinguishes between classes (such as canines versus felines). Knowing that within the class \textcolor{violet}{\textit{Canines}} there are subclasses such as \textcolor{violet}{\textit{Cocker}}, \textcolor{violet}{\textit{Labrador}}, and \textcolor{violet}{\textit{Shih Tzu}}, it is essential that all these subclasses are represented in the ID training folds. Otherwise, the model may learn only from some subclasses (for example, \textcolor{violet}{\textit{Cocker}} and \textcolor{violet}{\textit{Labrador}}), and when encountering an unseen subclass like \textcolor{violet}{\textit{Shih Tzu}}, it may incorrectly classify it as OOD, even though it belongs to the same class and superclass. This occurs because the model was never exposed to \textcolor{violet}{\textit{Shih Tzu}} in the ID folds and thus failed to learn to recognize it due to incorrect sampling, as illustrated in Figure~\ref{ejem_error}. This example highlights a broader issue: the greater the semantic similarity between the classes to be distinguished, the less reliable it is to rely on the generalization ability of the model.

\subsubsection*{DCV-ROOD in datasets with class hierarchy}
The application of the DCV-ROOD framework to hierarchical classes, similarly to non-hierarchical scenarios, involves defining two main splits. One is a k-fold CV split for the ID data, and the other is a k-fold CV split for the OOD data. The hierarchy introduces the need to identify the classification level: the semantic level at which the classification model is trained and through which an instance is determined to be ID or OOD. No class at this level should appear in both partitions, as this would imply that the same data could be considered both ID and OOD, generating inconsistencies in the evaluation.

The first step consists of creating the ID and OOD splits, ensuring their independence from the classes at the classification level. Classes from the level immediately above the classification level are treated as strata. Within each stratum, several classes at the classification level are selected to be OOD. If the dataset is balanced, a fixed number of classes per stratum can be chosen, if it is imbalanced, a proportional percentage is selected.

The procedure ensures that both ID and OOD splits include classes from all levels above the classification level. This prevents OOD detection from being artificially simplified due to the ID and OOD splits representing a greater semantic difference than actually exists. It also guarantees that classes at the classification level appear in only one of the splits (ID or OOD), avoiding duplication. Duplication would imply that the same data could be considered both ID and OOD, generating inconsistencies in the evaluation. In some cases, the selection of ID and OOD classes may be determined by the nature of the dataset or by experimental design constraints. For example, classes with insufficient sample size could be assigned as OOD, or certain ID classes may be excluded for reasons related to the objectives of the analysis.

After defining the ID and OOD partitions, the second step is to generate CV folds for each split.
\begin{itemize}
    \item For ID classes:
in the ID split, a stratified k-fold is applied to the last level of the hierarchy, whether or not it corresponds to the classification level. This ensures that all classes from all levels are represented in each fold, including levels beyond the classification level.

     \item For OOD classes:
in the OOD split, group k-fold is iteratively applied at the classification level within each class of the level immediately above. The resulting folds are then merged at the fold level. This guarantees that each OOD class at the classification level appears in only one fold and that all higher levels are represented in every fold. This ensures that during training with outlier exposure, test OOD classes never appear in the training data. By stratifying at the preceding level, all higher levels of the hierarchy are represented in each fold, providing a more representative and coherent test with respect to the dataset’s class hierarchy.
\end{itemize}

\begin{figure}[H]
    \makebox[\textwidth][c]{%
        \includegraphics[scale=0.33]{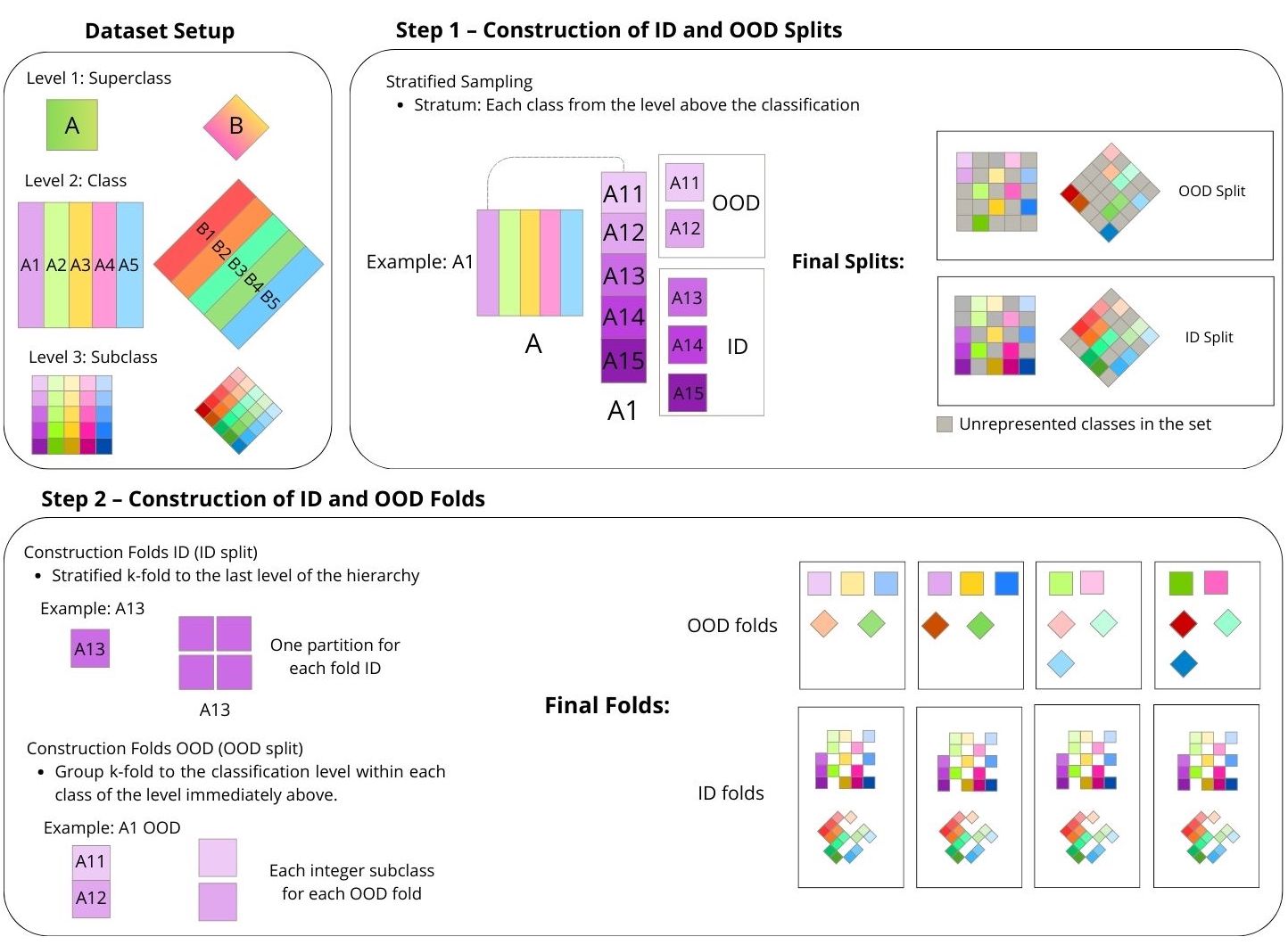}%
    }
    \caption{Balanced hierarchical classes with 2 superclasses, each containing 5 classes and 5 subclasses per class. Classification is performed at the subclass level under DCV-ROOD (4-fold), with 60\% of subclasses designated as ID and 40\% as OOD.}
    \label{new}
\end{figure}

The procedure ensures a proper balance between ID and OOD classes, maintains hierarchical consistency across all partitions, and provides consistent and robust evaluations for OOD detection in hierarchical settings. In this way, the proposed methodology prevents information leakage and preserves the integrity of the hierarchy at all stages of sampling. Figure~\ref{new} provides an example of applying DC-ROOD to a hierarchical classes, illustrating the approach in practice.

\subsection{Algorithms for DCV-ROOD framework}

In this section we describe the proposed algorithms for the DCV-ROOD framework. We distinguish between the two previously described scenarios: the baseline case, non-hierarchical classes, and the case with hierarchical classes, where it becomes necessary to carefully define the OOD partitions in order to ensure a fairer evaluation. Several parameters will be defined for the algorithms, whose notation and description can be found in Table \ref{tbl:algparams}.

\begin{table}[H]
\centering
\caption{Description of the parameters used in the algorithms}
\label{tbl:algparams}
\begin{tabular}{ll}
\toprule
Item & Description \\
\midrule
\multicolumn{2}{l}{\textbf{Class-level hierarchy}} \\[0.5ex]
$\{ \text{Level 1} \to \dots \to S \to C \to \dots \to L \}$ &  $2 \le |\{\text{Levels}\}| < \infty$ (for a dataset with class hierarchy) \\[1ex]
$\{C\}$ & (for a dataset without class hierarchy) \\[1ex]
\midrule
\multicolumn{2}{l}{\textbf{Definitions}} \\[0.5ex]
$S$ & Level preceding the classification. \\[1ex]
$C$ & Classification level. \\[1ex]
$L$ & Deepest hierarchical level available. \\[1ex]
$N$ & Total number of classes at the hierarchical level $S$, referred to as strata. \\[1ex]
$S_i$ & $i$-th class at level $S$, $i = 1, \ldots, N$. \\[1ex]
$N_i$ & Number of subclasses within $S_i$. \\[1ex]
$C_{ij}$ & $j$-th subclass of $S_i$, $j = 1, \ldots, N_i$. \\[1ex]
\midrule
\multicolumn{2}{l}{\textbf{CV procedures}} \\[0.5ex]
$\textsc{stratifiedKFold}(\texttt{dataset},\ \texttt{stratas},\ \texttt{K})$ & 
Stratified $K$-fold using the specified strata of the dataset. \\[1ex]

$\textsc{groupKFold}(\texttt{dataset},\ \texttt{group\_by},\ \texttt{K})$ & 
Group $K$-fold keeping each group indivisible in the dataset. \\[1ex]
\multicolumn{2}{l}{\textbf{Parameters}} \\[0.5ex]
\texttt{dataset} & A labeled (both with or without class hierarchy) dataset to split. \\[1ex]
\texttt{stratas} / \texttt{group\_by} & The specific level in the class hierarchy to be used to make the folds. \\[1ex]
\texttt{$K$} & Number of folds to obtain. \\[1ex]

\midrule
\multicolumn{2}{l}{\textbf{Other procedures}} \\[0.5ex]
\textsc{sampleWithoutReplacement}($\Omega$, $n$) & Sample without replacement $n$ items from the set $\Omega$. \\[1ex]
\textsc{filterDatasetByLabels}($\mathcal{H}$, $\mathcal{C}$) & Get the subset of a dataset $\mathcal{H}$ with labels belonging to a set of labels $\mathcal{C}$. \\[1ex]
\textsc{joinByFold}($\{\mathcal{F}_1, \dots, \mathcal{F}_t\})$ & Create a single $k$-fold by aligning $k$-folds $\mathcal{F}_1, \dots, \mathcal{F}_t$ index-wise. \\[1ex] 
\bottomrule
\end{tabular}
\end{table}

\subsubsection*{Applying DCV-ROOD to non-hierarchical classes}

In the case without a class hierarchy, it is common to encounter scenarios where two different datasets are used, serving as ID and OOD respectively. If starting from a dataset without predefined ID and OOD splits, these can be established through simple random sampling over the dataset classes. Once the ID/OOD split is defined, the folds of DCV-ROOD can be generated for each subset accordingly, as described in Section \ref{ssec:cv_ood}. Algorithm~\ref{alg:no_jerarquico} describes this procedure. Here, we use stratified $k$-fold for the ID data and group $k$-fold for the OOD data since these are the more inclusive approaches, although any of the fold-partitioning methods described in Section~\ref{ssec:cv_ood} may be considered.




\begin{algorithm}[H]
\caption{Fold construction for non-hierarchical classes}
\label{alg:no_jerarquico}

\begin{algorithmic}[1]
    \State \textbf{Input:} Labeled ID dataset $\mathcal{D}^\text{ID}$, Labeled OOD dataset $\mathcal{D}^{\text{OOD}}$, Number of folds $K$
    \State \textbf{Output:} ID Folds $\mathcal{F}^{\text{ID}}$, OOD folds $\mathcal{F}^{\text{OOD}}$
    \State $\mathcal{F}^{\text{ID}} \gets \textsc{stratifiedKFold}(\mathcal{D}^\text{ID}, C, K)$
    \State $\mathcal{F}^{\text{OOD}} \gets \textsc{groupKFold}(\mathcal{D}^{\text{OOD}}, C, K)$
    \State \Return $\mathcal{F}^{\text{ID}},\ \mathcal{F}^{\text{OOD}}$
\end{algorithmic}

\end{algorithm}

\subsubsection*{Applying DCV-ROOD to hierarchical classes}

As shown in subsection~\ref{hierarchical}, hierarchical classes require special considerations for proper splitting. On one hand, if ID and OOD splits are not predefined, they must be generated (see Algorithm~\ref{alg_1}). On the other hand, once the ID and OOD splits are defined, the corresponding folds for CV must be created (see Algorithm~\ref{alg_2}).

Algorithm~\ref{alg_1} is applied to hierarchical classes without a predefined ID/OOD split. Its purpose is to produce two splits, one for ID classes and another for OOD classes, reflecting the variability and semantic complexity of the dataset. To achieve this, it performs stratified sampling within each stratum, assigning a specified percentage of subclasses from each stratum to OOD and the remainder to ID, using the classes at the level immediately above the classification level as strata.


\begin{algorithm}
    \caption{Stratified selection into ID and OOD splits}
    \label{alg_1}
    \hspace*{\algorithmicindent} \textbf{Input}: Dataset with class hierarchy $\mathcal{H}$\\
    \hspace*{\algorithmicindent} \textbf{Input}: Strata $\{S_i\}_{i=1}^N$, with $S_i = \{C_{i1}, \ldots, C_{iN_i}\}$\\
    \hspace*{\algorithmicindent} \textbf{Input}: Proportion $p \in [0,1]$ of subclasses assigned to OOD\\
    \hspace*{\algorithmicindent} \textbf{Output}: ID split $\mathcal{H}^{\text{ID}}$, OOD split $\mathcal{H}^{\text{OOD}}$
    \begin{algorithmic}[1]
        \State $\mathcal{C}^{\text{ID}} \gets \emptyset$, $\mathcal{C}^{\text{OOD}} \gets \emptyset$
        \For{$i = 1,\ldots,N$}
            \State $k_i \gets \lfloor p \cdot N_i \rfloor$ \Comment{Integer number of subclasses from $S_i$ assigned to OOD}
            \State $O_i \gets \textsc{sampleWithoutReplacement}(S_i, k_i)$ \Comment{Random subset of size $k_i$}
            \State $I_i \gets S_i \setminus O_i$ \Comment{Remaining subclasses for ID}
            \State $\mathcal{C}^{\text{OOD}} \gets \mathcal{C}^{\text{OOD}} \cup O_i$ \Comment{Append all OOD classes}
            \State $\mathcal{C}^{\text{ID}} \gets \mathcal{C}^{\text{ID}} \cup I_i$ \Comment{Append all ID classes}
        \EndFor
        \State $\mathcal{H}^{\text{ID}} \gets \textsc{filterDatasetByLabels}(\mathcal{H}, \mathcal{C}^{\text{ID}})$ \Comment{Subset of $\mathcal{H}$ with labels belonging to $\mathcal{C}^{\text{ID}}$}
        \State $\mathcal{H}^{\text{OOD}} \gets \textsc{filterDatasetByLabels}(\mathcal{H}, \mathcal{C}^{\text{OOD}})$ \Comment{Subset of $\mathcal{H}$ with labels belonging to $\mathcal{C}^{\text{OOD}}$}
        \State \Return $\mathcal{H}^{\text{ID}}$, $\mathcal{H}^{\text{OOD}}$ splits
    \end{algorithmic}
\end{algorithm}

Algorithm~\ref{alg_2} is used once the ID and OOD partitions are established. It begins by applying Algorithm~\ref{alg_1} to obtain the initial ID/OOD split, after which Algorithm~\ref{alg_2} constructs the $k$-folds for ID and OOD data using this initial split. Specifically, it applies a stratified k-fold to the classes at the lowest hierarchical level in the ID split, and a group k-fold to the OOD split at the level immediately above the classification level.


\begin{algorithm}
    \caption{Creating k-Folds for ID and OOD splits}
    \label{alg_2}
    \hspace*{\algorithmicindent} \textbf{Input}: ID split $\mathcal{H}^{\text{ID}}$ and OOD split $\mathcal{H}^{\text{OOD}}$ of a dataset with class hierarchy \\
    \hspace*{\algorithmicindent} \textbf{Input}: Number of folds $K$\\
    \hspace*{\algorithmicindent} \textbf{Output}: ID folds $\mathcal{F}^{\text{ID}}$, OOD folds $\mathcal{F}^{\text{OOD}}$
    \begin{algorithmic}[1]
        \State $\mathcal{F}^{\text{ID}}$ $\gets$ \textsc{stratifiedKFold}($\mathcal{H}^{\text{ID}}$, $L$, $K$) \Comment{Stratified $k$-fold over the deepest level in the hierarchy for ID}
        \For{$i \gets 1$ to $N$}
            \State $\mathcal{H}^{\text{OOD}}_i \gets \textsc{filterDatasetByLabels}(\mathcal{H}^{\text{OOD}}, S_i)$ \Comment{Subset of $\mathcal{H}^{\text{OOD}}$ belonging to $S_i$} 
            \State $\mathcal{F}^{\text{OOD}}_i \gets \textsc{groupKFold}(\mathcal{H}^{\text{OOD}}_i, C, K)$ \Comment{Group $k$-fold on the class level for OOD, for each $S_i$}
        \EndFor
        \State $\mathcal{F}^{\text{OOD}} \gets \textsc{joinByFold}\left( \{\mathcal{F}^{\text{OOD}}_1, \dots, \mathcal{F}^{\text{OOD}}_N \} \right)$
        \State \Return $\mathcal{F}^{\text{ID}}, \mathcal{F}^{\text{OOD}}$
    \end{algorithmic}
\end{algorithm}

Together, Algorithm~\ref{alg_1} handles the initial generation of the ID/OOD split, while Algorithm~\ref{alg_2} constructs representative CV folds for evaluation. If the ID and OOD splits are already provided, the procedure can proceed directly with Algorithm~\ref{alg_2} without executing Algorithm~\ref{alg_1}.

A series of experiments are presented below, conducted to assess the effectiveness of the proposed DCV-ROOD framework and to examine the performance of various OOD image detection methods.

\section{Experimental Framework}\label{sec:experiments}
This section outlines the experimental setup designed to evaluate the proposed DCV-ROOD evaluation framework for OOD detection. In Section~\ref{datasets}, we describe the datasets used, while the OOD detection methods employed are described in Section~\ref{methods}, the evaluation metrics applied are shown in Section~\ref{metrics} and the training process is illustrated in Section~\ref{training}. The experimental studies conducted and the results obtained, along with the analysis of the effectiveness of the proposed DCV-ROOD approach are discussed in Section~\ref{sec:analysis}.

\subsection{Datasets}\label{datasets}

The following section provides a concise description of the datasets employed in the experiments.

\begin{itemize}
 
\item \textbf{CIFAR.} The two CIFAR datasets~\cite{netzer2011reading} consist of color images with a resolution of 32×32 pixels. The CIFAR10 dataset consists of 60000 32x32 color images in 10 classes, with 6000 images per class. There are 50000 training images and 10000 test images. CIFAR100 has 100 classes containing 600 images each, 500 training images and 100 testing images per class. The 100 classes in the CIFAR100 are grouped into 20 superclasses.

\item \textbf{Tiny ImageNet.} Tiny ImageNet~\cite{le2015tiny} is a downsized subset of the ImageNet dataset~\cite{russakovsky2015imagenet}, consisting of color images with a resolution of 64x64 pixels. It includes 200 classes, each containing 500 training images, 50 validation images, and 50 test images, totaling 120000 images.

\item \textbf{DTD.} The Describable Textures Dataset (DTD)~\cite{cimpoi2014describing} contains 5640 color images categorized into 47 describable texture attributes, such as \emph{striped}, emph{rough}, or emph{woven}. Each class contains 120 images. The images vary in resolution and were collected under uncontrolled conditions, making the dataset suitable for studying texture recognition in the wild.

\item \textbf{EMNIST.} The EMNIST dataset~\cite{cohen2017emnist} is an extension of the original MNIST dataset, introducing several new partitions. In this work, we use only the Letters and MNIST splits. EMNIST Letters comprises 145600 training images and 14500 test images distributed across 26 classes, representing the uppercase letters of the English alphabet without case distinction. EMNIST MNIST replicates the original digit dataset, consisting of grayscale 28x28 pixel images of handwritten digits (0–9), with 60000 training samples and 10000 test samples.

\end{itemize}

Given these datasets, we created ID-OOD pairs for experimentation. Table~\ref{tab:ID_OOD} presents the resulting pairs. Notably, SUPERCIFAR100-ID vs. SUPERCIFAR100-OOD was generated by first applying Algorithm~\ref{alg_1} to obtain the ID/OOD split, followed by Algorithm~\ref{alg_2} to construct the folds, using the CIFAR100 dataset.

\begin{table}[h]
\centering
\caption{ID vs. OOD dataset}
\label{tab:ID_OOD}
\begin{tabular}{|c|c|}
\hline
\textbf{Near} & \textbf{Far} \\
\hline
CIFAR10 vs. CIFAR100 & CIFAR10 vs. DTD \\
CIFAR10 vs. TinyImageNet & CIFAR10 vs. MNIST \\
CIFAR100 vs. CIFAR10 & CIFAR10 vs. LETTERS \\
CIFAR100 vs. TinyImageNet & CIFAR100 vs. DTD \\
SUPERCIFAR100-ID vs. SUPERCIFAR100-OOD & CIFAR100 vs. MNIST \\
 & CIFAR100 vs. LETTERS \\
\hline
\end{tabular}
\end{table}


\subsection{OOD detection methods}\label{methods}

In this section, we briefly describe the OOD detection methods employed in this study. The implementations were obtained from the OpenOOD repository~\cite{zhang2023openood}. A categorization of the methods is provided in Table~\ref{tab:OODmethods}.

\begin{itemize} 

\item \textbf{KNN.} The K-Nearest Neighbors (KNN) baseline~\cite{sun2022out} performs OOD detection by computing distances between a test sample’s feature embedding and those of training data.

\item \textbf{MDS.} The Mahalanobis Distance-based Score (MDS) method~\cite{NEURIPS2018_abdeb6f5} models class-conditional feature distributions. It computes the Mahalanobis distance of a test sample to the nearest class mean in feature space.

\item \textbf{KLM.} The Kernel Log-Magnitude (KLM) method~\cite{hendrycks2022scaling} is a density-based OOD detection technique that estimates the likelihood of a sample in feature space using kernel density estimation. The detection score is computed from the log-magnitude of the estimated density.

\item \textbf{GEN.} The Generalized Entropy (GEN) method~\cite{liu2023gen} is a post-hoc OOD detection approach that leverages the predictive distribution of a pre-trained softmax classifier. Without requiring access to training data or model re-training, GEN computes an entropy-based score solely from the model's output probabilities.

\item \textbf{fDBD.} The Fast Decision Boundary based OOD Detector (fDBD)~\cite{liu2023fast} is a computationally efficient, post-hoc OOD detection method that estimates the distance of a sample's feature representation to the decision boundary. By deriving a closed-form lower bound for this distance, fDBD identifies that ID samples typically lie farther from the decision boundary compared to OOD samples.

\item \textbf{EBO.} The Energy-based model for OOD detection~\cite{liu2020energy} computes an energy score derived from the log-sum-exp of the model's logits. Unlike softmax confidence, the energy score captures the unnormalized log-likelihood of the input under the model. Lower energy values are typically associated with ID data, while higher energy indicates potential OOD samples.

\item \textbf{Relation.} The Relation method~\cite{kim2023neural} constructs a Neural Relation Graph to model sample-wise relational structures in the feature space. By learning the similarity relationships among samples, it captures contextual dependencies and distinguishes outlier or mislabeled data based on inconsistencies within this relational graph. The framework unifies the identification of label noise and OOD instances, enabling robust detection in scenarios with complex data irregularities.

\item \textbf{NNGuide.} Nearest Neighbor Guidance (NNGuide)~\cite{park2023nearest} is a post-hoc OOD detection method that adjusts classifier confidence scores based on the similarity between test inputs and their nearest neighbors in the training set. By guiding the classifier's output to respect the boundary geometry of the data manifold, NNGuide reduces overconfidence in regions far from the training data, thereby enhancing the detection of OOD samples.

\end{itemize}

\begin{table}[h]
\centering
\caption{\label{tab:OODmethods}OOD Detection Methods}
\begin{tabular}{|c|c|c|}
\hline
\textbf{Density} & \textbf{Classification} & \textbf{Distance} \\
\hline
EBO              & GEN                    & MDS               \\
KLM              & NNGuide                & fDBD              \\
                 & KNN                    &                   \\
                 & Relation               &                   \\
\hline
\end{tabular}
\end{table}


\subsection{OOD evaluation metrics}\label{metrics}

We evaluate OOD detection methods based on their capability to reliably distinguish OOD samples from ID ones. In this context, OOD examples are treated as the positive class. To comprehensively assess performance, we consider five evaluation metrics: True Positive Rate at N\% False Positive Rate (TPRN), Area Under the Receiver Operating Characteristic Curve (AUROC), Area Under the Precision-Recall Curve (AUPR), F1-score  and overall Accuracy.

\begin{itemize}

\item  \textbf{TPRN.} TPRN, conversely, quantifies the proportion of OOD samples correctly detected when the false positive rate is fixed at N\%. This metric highlights how well a detector performs under a controlled level of false alarms. We report TPR5, corresponding to a false positive rate fixed at 5\%.

\item  \textbf{AUROC.} AUROC provides a threshold-independent measure of a detector’s ability to rank OOD samples above ID samples. It can be interpreted as the probability that a randomly chosen OOD sample receives a higher anomaly score than a randomly chosen ID sample~\cite{davis2006relationship}. A perfect detector achieves an AUROC of 100\%, while an uninformative one scores 50\%.

\item  \textbf{AUPR.} AUPR focuses on the precision-recall trade-off, which is particularly informative in imbalanced settings where OOD samples are rare. It summarizes the detector’s performance across varying thresholds, emphasizing its ability to maintain high precision while achieving high recall.

\item \textbf{F1-score.} The F1-score is a metric that combines two key measures, precision and recall, by summarizing them into a single value using the harmonic mean. For a more in-depth discussion, we refer the reader to \cite{christen2023review}.

\item  \textbf{Accuracy.} Accuracy represents the overall correctness of OOD vs. ID classification based on a fixed threshold, typically chosen using a validation set. While it is threshold-dependent, it offers an intuitive sense of the detector’s classification reliability. In our experiments, we report accuracy at thresholds corresponding to 90\%.

\end{itemize}
Together, these metrics provide a balanced and nuanced evaluation of OOD detection performance across both threshold-sensitive and threshold-independent criteria.

\subsection{Training of the models}\label{training}

A ViT-B16 transformer architecture is trained for classification on the ID dataset, using weights pretrained on ImageNet1K\_V1. The classification head is adapted to match the number of ID classes. Training is carried out with early stopping based on validation performance, using a patience of 3 epochs. After identifying the best-performing epoch, we evaluate the selected OOD detection methods on the corresponding test set.

Subsequently, the model is fine-tuned using Outlier Exposure (OE). This method involves incorporating outlier examples during training. The loss function for outliers is designed to flatten their logits, encouraging uniform predictions across all ID classes. In the work by Hendrycks et al. ~\cite{hendrycks2018deep}, applying OE resulted in improved OOD detection performance across multiple datasets   and metrics. In our setup, we incorporate the training split of the OOD dataset for an additional 5 epochs. The OOD detection methods are then re-evaluated to assess the impact of OE on detection performance.

This training methodology will be consistently applied at the partition level, both for the training partitions used to compute the benchmark truth described in Section \ref{sec:bt} and for the training partitions corresponding to each fold generated by the DCV-ROOD framework.

\section{Experimental Study}\label{sec:analysis}

This section presents the studies conducted within the experimental framework described in Section \ref{sec:experiments} to assess the effectiveness of DCV-ROOD as an evaluation framework. First, Section \ref{sec:bt} outlines the procedure designed to construct a reference framework representing the expected generalization performance under ideal conditions, which we refer to as the \emph{benchmark truth}. Next, two distinct studies are conducted to compare the partitions generated by DCV-ROOD with those from the benchmark truth. Both analyses focus on assessing the similarity between the results obtained from the benchmark truth and those derived from DCV-ROOD. The first study (Section \ref{sec:study1}) compares the number of statistically significant differences among OOD detection methods identified in the benchmark truth with those observed across each DCV-ROOD instance. The second study (Section \ref{sec:study2}) examines whether significant differences exist between the performance of each method under the benchmark truth and its corresponding performance under DCV-ROOD, applying this comparison consistently across all evaluated methods.

\subsection{Benchmark truth}\label{sec:bt}
In order to evaluate the effectiveness of the proposal, a reliable benchmark is essential. This serves as a form of \emph{reference truth} for comparing the results. Since the focus is on models for detecting OOD data, the benchmark truth refers to the performance of the evaluated methods. The evaluation setup is similar to that used in~\cite{moreno2012study}, although adapted to the specific objectives of the present study.

To ensure objectivity in building the benchmark truth, the convergence of results across different datasets was analyzed. Sampling was adapted to each ID/OOD split. For ID data, simple random sampling was applied, as their class distributions are balanced. A subset of OOD classes was reserved when OOD data was used during training. Additionally, separate ID data and exclusive OOD classes were set aside for testing. This procedure was repeated 100 times using different random seeds.

Convergence was analyzed using the mean convergence method. A setup based on 100 repetitions was found to be sufficient to conclude that the results had converged. Therefore, the mean of these 100 experiments is considered a reliable estimate of the benchmark truth. Table~\ref{gran_verdad} presents the results for the TPR5 metric, one of the reference metrics used in this study. For completeness, results corresponding to other evaluation metrics are included in Appendix~\ref{Anexo}. Based on this reference, the accuracy and efficiency of the proposed DCV-ROOD scheme were evaluated.

\begin{table}[ht]
\centering
\caption{Benchmark Truth (TPR5 on OOD Detection, 100 Iterations to Convergence)}
\label{gran_verdad}

\begin{adjustbox}{width=\textwidth}
\begin{tabular}{lllcccccccc}
\toprule
versions & id\_database & ood\_database & ebo & fdbd & gen & klm & knn & mds & nnguide & relation \\
\midrule
with ood         & cifar10          & cifar100           & 0.7172 & 0.6828 & 0.7175 & 0.1102 & 0.7145 & 0.6820 & 0.6500 & 0.6537 \\
                 &                  & dtd                & 0.9970 & 0.9963 & 0.9979 & 0.9888 & 0.9898 & 0.9924 & 0.9986 & 0.9946 \\
                 &                  & letters            & 0.9985 & 0.9984 & 0.9985 & 0.9970 & 0.9977 & 0.9978 & 0.9986 & 0.9985 \\
                 &                  & mnist              & 1.0000 & 1.0000 & 1.0000 & 0.9995 & 1.0000 & 0.9998 & 1.0000 & 1.0000 \\
                 &                  & tiny\_imagenet\_200 & 0.8681 & 0.8231 & 0.8725 & 0.6652 & 0.8671 & 0.9028 & 0.8119 & 0.7806 \\
                 & cifar100         & cifar10            & 0.2667 & 0.2047 & 0.2709 & 0.1170 & 0.1758 & 0.2911 & 0.2097 & 0.2593 \\
                 &                  & dtd                & 0.9681 & 1.0000 & 0.9999 & 0.9885 & 0.9854 & 0.9946 & 1.0000 & 1.0000 \\
                 &                  & letters            & 0.9711 & 1.0000 & 1.0000 & 0.9934 & 0.9835 & 0.9928 & 1.0000 & 1.0000 \\
                 &                  & mnist              & 0.9705 & 1.0000 & 1.0000 & 0.9931 & 1.0000 & 1.0000 & 1.0000 & 1.0000 \\
                 &                  & tiny\_imagenet\_200 & 0.9699 & 1.0000 & 1.0000 & 0.9937 & 0.9964 & 0.9974 & 1.0000 & 1.0000 \\
                 & super\_cifar100-ID  & super\_cifar100-OOD    & 0.3586 & 0.3181 & 0.3543 & 0.0920 & 0.3596 & 0.4131 & 0.3097 & 0.2165 \\
without ood         & cifar10          & cifar100           & 0.6926 & 0.3588 & 0.6994 & 0.0472 & 0.7282 & 0.7719 & 0.4856 & 0.3919 \\
                 &                  & dtd                & 0.9482 & 0.7321 & 0.9439 & 0.6904 & 0.9445 & 0.9969 & 0.9365 & 0.8535 \\
                 &                  & letters            & 0.7798 & 0.4372 & 0.7252 & 0.1605 & 0.7797 & 0.8702 & 0.7960 & 0.6522 \\
                 &                  & mnist              & 0.8231 & 0.5784 & 0.7992 & 0.1442 & 0.8297 & 0.9013 & 0.7713 & 0.6684 \\
                 &                  & tiny\_imagenet\_200 & 0.6068 & 0.3390 & 0.6177 & 0.0346 & 0.6521 & 0.7748 & 0.4395 & 0.3515 \\
                 & cifar100         & cifar10            & 0.1497 & 0.1825 & 0.1657 & 0.0192 & 0.1839 & 0.4122 & 0.0986 & 0.2142 \\
                 &                  & dtd                & 0.5741 & 0.2083 & 0.5588 & 0.0803 & 0.4961 & 0.7485 & 0.4339 & 0.2669 \\
                 &                  & letters            & 0.1841 & 0.2960 & 0.2273 & 0.1093 & 0.2624 & 0.2487 & 0.3635 & 0.3483 \\
                 &                  & mnist              & 0.1636 & 0.2780 & 0.2033 & 0.1087 & 0.2348 & 0.1901 & 0.3334 & 0.3252 \\
                 &                  & tiny\_imagenet\_200 & 0.3304 & 0.1823 & 0.3176 & 0.1122 & 0.3546 & 0.4631 & 0.2385 & 0.1212 \\
                 & super\_cifar100-ID  & super\_cifar100-OOD    & 0.3408 & 0.2822 & 0.3451 & 0.0910 & 0.3359 & 0.4011 & 0.2858 & 0.1507 \\
\bottomrule
\end{tabular}
\end{adjustbox}
\end{table}





\subsection{Overall statistical comparison between DCV-ROOD and benchmark truth}
\label{resultados_exp} \label{sec:study1}

This section evaluates the fidelity of the DCV-ROOD validation framework by comparing its results with the  benchmark truth. The analysis investigates the degree of agreement between statistically significant differences in OOD detection methods observed in the benchmark truth and those detected by DCV-ROOD across 10 different CV experiments.

After reviewing several studies analyzing CV by comparing the results with a benchmark truth~\cite{moreno2012study, raeder2010consequences}, we considered it appropriate to adopt a similar strategy to carry out an initial comparison. This approach allows us to evaluate the extent to which DCV-ROOD preserves the statistical relationships observed in the benchmark truth and, consequently, its reliability as a validation technique.

The first step involves identifying which OOD detection methods show statistically significant differences according to the benchmark truth results, and quantifying how often this pattern is repeated in DCV-ROOD. The more such matches occur, the greater the fidelity of the DCV-ROOD results to the benchmark truth.

To quantify the \emph{fidelity to the benchmark truth}, we define a \emph{match} as a case in which a statistically significant difference between two methods is detected in the DCV-ROOD experiments, and the same difference is also present in the benchmark truth. We will then analyze the results to assess the degree of agreement between the proposed DCV-ROOD and the benchmark truth.

To analyze the statistically significant differences between methods, a normality test was first conducted. In general, the results of the methods did not satisfy the normality assumption across the different metrics evaluated throughout the 10 CV experiments. The only cases in which the results exhibited behavior consistent with a normal distribution were the \textit{gen} and \textit{relation} methods in the fifth CV experiment.

Since pairwise comparisons were conducted to identify significant differences between methods in each CV experiment, and given that the normality assumption was not met in most cases, non-parametric statistical tests were used. Specifically, the Mann–Whitney rank test (also known as the Mann–Whitney U test) was employed. Unlike the other experiments, in the fifth CV experiment the Mann–Whitney U test was used for all pairwise method comparisons except for \textit{gen} and \textit{relation}, for which, after confirming normality in both methods, the Student’s \textit{t}-test was applied.

Table~\ref{test_verdad} presents the results of the Mann–Whitney rank test for the TPR5 metric, where the null hypothesis states that there are no statistically significant differences between the pairs of compared methods.

\begin{table}[ht]
\centering
\begin{threeparttable}
\caption{Significant Differences in the Benchmark Truth Based for the TPR5 Metric}
\label{test_verdad}
\begin{tabular}{lcccccccc}
\toprule
 & ebo & fdbd & gen & klm & knn & mds & nnguide & relation \\
\midrule
ebo      & 1          & \textbf{0.0301**}  & 0.1289       & \textbf{0.0003***} & 0.1560      & \textbf{0.0001***} & 0.1560      & \textbf{0.0738*} \\
fdbd     & \textbf{0.0301**} & 1          & \textbf{0.0141***} & \textbf{0.0000***} & \textbf{0.0462**} & \textbf{0.0229***} & \textbf{0.0157***} & 0.4628 \\
gen      & 0.1289     & \textbf{0.0141***} & 1          & \textbf{0.0000***} & 0.6327      & \textbf{0.0093***} & 0.1659      & \textbf{0.0042***} \\
klm      & \textbf{0.0003***} & \textbf{0.0000***} & \textbf{0.0000***} & 1          & \textbf{0.0000***} & \textbf{0.0000***} & \textbf{0.0000***} & \textbf{0.0000***} \\
knn      & 0.1560     & \textbf{0.0462**} & 0.6327      & \textbf{0.0000***} & 1          & \textbf{0.0017***} & 0.2099      & \textbf{0.0587*} \\
mds      & \textbf{0.0001***} & \textbf{0.0229***} & \textbf{0.0093***} & \textbf{0.0000***} & \textbf{0.0017***} & 1          & \textbf{0.0275***} & \textbf{0.0115***} \\
nnguide  & 0.1560     & \textbf{0.0157***} & 0.1659      & \textbf{0.0000***} & 0.2099      & \textbf{0.0275***} & 1          & \textbf{0.0025***} \\
relation & \textbf{0.0738*} & 0.4628     & \textbf{0.0042***} & \textbf{0.0000***} & \textbf{0.0587*} & \textbf{0.0115***} & \textbf{0.0025***} & 1          \\
\bottomrule
\end{tabular}
\begin{tablenotes}
\footnotesize
\item \textbf{Note:} Significance levels are indicated as follows: * $p < 0.1$, ** $p < 0.05$, *** $p < 0.01$.
\end{tablenotes}
\end{threeparttable}
\end{table}

Table~\ref{aciertos} shows the number of times each pair of methods exhibited statistically significant differences across the DCV-ROOD experiments. Since 10 DCV-ROOD runs were performed, the maximum possible value for each method pair is 10. For example, a value of 9 between the methods Relation and KLN indicates that a statistically significant difference was observed between them in 9 out of the 10 DCV-ROOD experiments. Bold values highlight those method pairs that also showed significant differences in the benchmark truth results.

\begin{table}[h]
\centering
\begin{threeparttable}
\caption{Hit Rate for the Metric TPR5\_pval$\leq$0.05}
\label{aciertos}
\begin{tabular}{l|cccccccc}
\textbf{} & \textbf{ebo} & \textbf{fdbd} & \textbf{gen} & \textbf{klm} & \textbf{knn} & \textbf{mds} & \textbf{nnguide} & \textbf{relation} \\
\hline
\textbf{ebo}      & - & \textbf{10} & 0 & \textbf{10} & 0 & \textbf{10} & -1 & \textbf{6} \\
\textbf{fdbd}     & \textbf{10} & - & \textbf{10} & \textbf{10} & \textbf{10} & \textbf{10} & \textbf{10} & 0 \\
\textbf{gen}      & 0 & \textbf{10} & - & \textbf{10} & 0 & \textbf{10} & -2 & \textbf{9} \\
\textbf{klm}      & \textbf{10} & \textbf{10} & \textbf{10} & - & \textbf{10} & \textbf{10} & \textbf{10} & \textbf{10} \\
\textbf{knn}      & 0 & \textbf{10} & 0 & \textbf{10} & - & \textbf{9} & 0 & -9 \\
\textbf{mds}      & \textbf{10} & \textbf{10} & \textbf{10} & \textbf{10} & \textbf{9} & - & \textbf{9} & \textbf{10} \\
\textbf{nnguide}  & \textbf{1} & \textbf{10} & \textbf{2} & \textbf{10} & 0 & \textbf{9} & - & \textbf{10} \\
\textbf{relation} & \textbf{6} & 0 & \textbf{9} & \textbf{10} & -9 & \textbf{10} & \textbf{10} & - \\
\end{tabular}
\begin{tablenotes}
\footnotesize
\item \textbf{Note:} Bold values indicate method pairs with significant differences also found in the benchmark truth. Negative signs denote significant differences found in DCV-ROOD but not in the benchmark truth.
\end{tablenotes}
\end{threeparttable}

\end{table}
Tables~\ref{tasa-acierto-error-01} and~\ref{tasa-acierto-error-05} present the average number of correct detections in DCV-ROOD, considering only the method pairs that showed statistically significant differences in the benchmark truth. Specifically, it reports the mean number of DCV-ROOD experiments (out of a maximum of 10) in which a significant difference was also detected for those same pairs. A value of 10 indicates that, in all cases where the benchmark truth revealed a significant difference between two methods, this difference was also observed in all 10 corresponding DCV-ROOD runs.

To illustrate more clearly the meaning of the values presented in Tables~\ref{tasa-acierto-error-01} and~\ref{tasa-acierto-error-05}, consider the following example: if, from Table~\ref{aciertos}, one sums the number of times DCV detected significant differences for each pair of methods that also showed a significant difference in the benchmark truth, and divides this sum by the total number of such pairs, the resulting value corresponds to the one reported in the table for the respective metric. This value represents, on average, how many times DCV-ROOD successfully replicated the differences identified in the benchmark truth, thus providing a measure of the proposed method’s ability to approximate the reference results. In the same manner, the error rate is computed by averaging the number of times CV reported a significant difference between method pairs that did not show a significant difference in the benchmark truth. It also ranges from 0 to 10 and works like the hit rate but in reverse, with lower values indicating fewer false detections. This provides an estimate of the frequency with which CV falsely detects differences not supported by the reference data.

It is worth noting that interpreting these cases as errors is, to some extent, relative. One could also argue that DCV-ROOD was able to detect significant differences that were not identified in the benchmark truth, possibly due to the limited number of replications. This is mentioned here merely as a reflection, which could be further explored through additional experiments if deemed relevant.

\begin{table}[h]
\centering
\begin{minipage}[t]{0.48\textwidth}
\centering
\caption{Hit rate and error rate by metric}
\label{tasa-acierto-error-01}
\begin{tabular}{l|c|c}
\textbf{Metric}              & \textbf{Hit rate} & \textbf{Error rate} \\
\hline
TPR5\_pval$\leq$0.1          & 9.8571  & 1.0000 \\
auroc\_pval$\leq$0.1         & 8.9130  & 1.600 \\
aupr\_pval$\leq$0.1          & 8.8636  & 0.5000 \\
F1\_score\_pval$\leq$0.1     & 8.9500  & 1.500 \\
acc\_90\_pval$\leq$0.1       & 8.6087  & 1.6000 \\
\end{tabular}
\end{minipage}
\hfill
\begin{minipage}[t]{0.48\textwidth}
\centering
\caption{Hit rate and error rate by metric}
\label{tasa-acierto-error-05}
\begin{tabular}{l|c|c}
\textbf{Metric}              & \textbf{Hit rate} & \textbf{Error rate} \\
\hline
TPR5\_pval$\leq$0.05          & 9.8421  & 2.0000 \\
auroc\_pval$\leq$0.05         & 9.2381  & 0.8571 \\
aupr\_pval$\leq$0.05          & 9.2778  & 1.7 \\
F1\_score\_pval$\leq$0.05     & 9.1176  & 2.0910 \\
acc\_90\_pval$\leq$0.05       & 8.0000  & 1.2857 \\
\end{tabular}
\end{minipage}
\end{table}


As shown in Tables~\ref{tasa-acierto-error-01} and~\ref{tasa-acierto-error-05}, the metric \texttt{TPR5\_pval$\leq$0.1} achieves the highest hit rate (9.8571), indicating that the significant differences observed in the benchmark truth were accurately replicated in almost all DCV-ROOD runs. Although its error rate (1.00) is higher than that of \texttt{AUPR\_pval$\leq$0.1}, which achieves the lowest error rate (0.50) with a hit rate of 8.8636, the error rate of \texttt{TPR5\_pval$\leq$0.1} remains relatively low.
On the other hand, considering a more stringent threshold (p$\leq$0.05), the metric \texttt{TPR5\_pval$\leq$0.05} achieves again the highest hit rate (9.8421), while \texttt{AUROC\_pval$\leq$0.05} has the lowest error rate (0.8571) with a hit rate of 9.2381, still reflecting a high value.

All metrics showed a hit rate above 8, indicating a high similarity between the DCV-ROOD results and those of the benchmark truth. Therefore, DCV-ROOD allowed for a robust prediction of model performance with a significantly lower computational cost compared to that required by the benchmark truth.

\subsection{Method-wise statistical comparison of DCV-ROOD and benchmark truth} \label{sec:study2}

In this section, the objective is also to evaluate the fidelity of the DCV-ROOD evaluation framework to the benchmark truth, from a different point of view. Specifically, statistically significant differences between OOD detection methods are analyzed by comparing each method’s results in the benchmark truth with the corresponding results of the same method across the same 10 DCV-ROOD iterations used in the previous study in Section~\ref{sec:study1}. That is, each method was compared a total of 10 times, corresponding to the results obtained in each DCV-ROOD iteration.

To this end, a non-parametric hypothesis test was performed, since none of the compared pairs simultaneously satisfied the normality assumption. Again, the Mann–Whitney rank test was employed to assess whether significant differences existed. The results for the evaluated metrics are presented in Table \ref{tab:significant-differences}. For each metric, the table shows the DCV-ROOD iterations in which statistically significant differences were found between at least one OOD detection method and its corresponding result in the benchmark truth. In cases where differences were found, the specific DCV-ROOD iteration and the methods that provided sufficient evidence to conclude their results differ from those in the benchmark truth are indicated.

\begin{table}[H]
\centering
\begin{threeparttable}
\caption{Significant differences between benchmark truth and DCV-ROOD}
\label{tab:significant-differences}
\begin{tabular}{l|c|c|c}
\toprule
\textbf{Metric} & \textbf{p $\leq$ 0.1} & \textbf{p $\leq$ 0.05} & \textbf{p $\leq$ 0.01} \\
\midrule
TPR5\_pval$\leq$0.05      & CV 3 (mds), CV 6 (mds) & CV 3 (mds)   & -               \\
auroc\_pval$\leq$0.05     & CV 3 (mds)             & -            & -               \\
aupr\_pval$\leq$0.1       & CV 3 (mds)             & -            & -               \\
F1\_score\_pval$\leq$0.1  & -                      & -            & -               \\
acc\_90\_pval$\leq$0.1    & -                      & -            & -               \\
\bottomrule
\end{tabular}
\begin{tablenotes}
\footnotesize
\item \textbf{Note:} Statistically significant differences between each OOD detection method’s results across independent DCV-ROOD iterations and their corresponding results in the benchmark truth. A dash (–) indicates no significant differences were found in any iteration.

\end{tablenotes}
\end{threeparttable}
\end{table}

Overall, the limited number of significant differences detected supports that the proposed DCV-ROOD evaluation framework for OOD detection can be a reliable approach to approximate the true performance of the methods without requiring an excessive number of experiments. DCV-ROOD proves to be robust as it maintains consistency in detecting significant differences across differents iterations, faithfully reflecting the results obtained with the benchmark truth. This indicates that the variability introduced by data partitioning did not adversely affect performance evaluation, probably due to the careful design of the DCV-ROOD evaluation framework.

\section{Conclusions}

\label{sec:conclusions}

This work presents DCV-ROOD, a evaluation framework designed for the robust evaluation of OOD detection methods. By integrating stratified k-fold CV for ID data with group k-fold strategies for OOD samples, DCV-ROOD effectively prevents data leakage and enables fair performance estimation across a wide range of OOD scenarios. Furthermore, this framework has been extended to accommodate a dataset with class hierarchy, incorporating class hierarchies to ensure realistic and challenging ID-OOD partitions.

Extensive experiments were conducted using state-of-the-art OOD detection techniques across multiple datasets and configurations, both with and without outlier exposure. A reference benchmark, based on 100 repetitions of randomized splits, was established to represent the true performance of the evaluated methods. The fidelity of DCV-ROOD was assessed by comparing the statistical differences it identifies with those found in the benchmark truth.

Results show that DCV-ROOD achieves high agreement with the benchmark truth in most evaluation metrics, consistently preserving significant differences between detection methods. Moreover, the framework maintains a low error rate, indicating that it introduces minimal spurious conclusions. These findings confirm that DCV-ROOD provides a statistically reliable and computationally efficient alternative to exhaustive experimentation, offering a robust framework for the evaluation of OOD detection systems in practical settings.

DCV-ROOD constitutes a significant step forward in the rigorous assessment of OOD detection systems. Its ability to faithfully replicate statistical distinctions observed under exhaustive evaluation, while drastically reducing computational cost, positions it as a robust and efficient alternative. Moreover, by mitigating the variability and biases associated with isolated train-test partitions, DCV-ROOD delivers more reliable and generalizable results. By addressing a critical gap in the evaluation of OOD detectors, DCV-ROOD strengthens the foundation for the development of trustworthy and responsible AI systems.

\subsection*{Acknowledgements}

This research results from the Strategic Project IAFER-Cib (C074/23), as a result of the collaboration agreement signed between the National Institute of Cybersecurity (INCIBE) and the University of Granada. This initiative is carried out within the framework of the Recovery, Transformation and Resilience Plan funds, financed by the European Union (Next Generation).

\bibliographystyle{splncs04}
\bibliography{bibliography.bib}

\clearpage

\appendix
\section{Appendix of results}
\label{Anexo}
The following presents the results obtained according to the metrics employed in this study: AUROC (Section~\ref{auroc}), AUPR (Section~\ref{aupr}), F1-score (Section~\ref{f1}), Accuracy@90 (Section~\ref{acc}) and TPR@5 (Section~\ref{tpr}).  Each subsection of this appendix presents up to four tables with the results according to these metrics, except for Section~\ref{tpr}, which includes only one table, as will be shown below.

The first table of each section (Tables~\ref{auroc_benchmark}, \ref{aupr_benchmark},~\ref{f1_benchmark}, and~\ref{accuracy_90}) display the Benchmark Truth results, obtained by repeating the experiment 100 times with different train-test splits. This analysis served to validate the stability of the detectors and to consider the results as the Benchmark Truth.

The second table of each section (Tables~\ref{test_verdad_auroc},~\ref{test_verdad_aupr},~\ref{test_verdad_f1}, and~\ref{test_verdad_acc}) present the significant differences in the Benchmark Truth, based on the results shown in the corresponding subsection for each metric.

The third table of each section (Tables~\ref{aciertos_auroc0.1},~\ref{aciertos_aupr}, \ref{aciertos_f1_01}, and~\ref{aciertos_acc90_01}) provide a comparative summary of statistically significant differences found between the methods evaluated via DCV-ROOD (10 experiments) and those found in the corresponding Benchmark Truth. Since none of the method pairs met the normality assumption, the non-parametric Mann–Whitney U test was used with a significance threshold of $p < 0.1$.

The fourth table of each section (Tables~\ref{aciertos_auroc0.5},~\ref{aciertos_aupr_005},~\ref{aciertos_f1_005}, and~\ref{aciertos_acc90_005}) present the same information as the third, but using a more stringent significance threshold of $p < 0.05$.

For the TPR5 metric, only one table is presented (Table~\ref{aciertos_tprfpr_01}), which shows the number of significant differences among OOD detection methods over the 10 DCV-ROOD experiments. The other tables related to this metric are located in subsection~\ref{resultados_exp} of the experiments: the Benchmark Truth table is Table~\ref{gran_verdad}, the table displaying significant differences for the Benchmark Truth is Table~\ref{test_verdad}, and the table referring to the Hit Rate for the TPR@FPR metric with p-value~$\leq$0.5 is Table~\ref{aciertos}.

\subsection{AUROC}
\label{auroc}

\begin{table}[H]
\centering
\caption{Benchmark Truth (AUROC on OOD Detection)}
\label{auroc_benchmark}
\begin{adjustbox}{width=\textwidth}
\begin{tabular}{lllcccccccc}
\toprule
versions & id\_database & ood\_database & ebo & fdbd & gen & klm & knn & mds & nnguide & relation \\
\midrule
with ood & cifar10 & cifar100 & 0.9206 & 0.9244 & 0.9316 & 0.8475 & 0.8692 & 0.8942 & 0.9185 & 0.9158 \\
& & dtd & 0.9988 & 0.9991 & 0.9996 & 0.9965 & 0.9930 & 0.9973 & 0.9996 & 0.9997 \\
& & letters & 0.9995 & 0.9996 & 0.9996 & 0.9988 & 0.9989 & 0.9991 & 0.9996 & 0.9998 \\
& & mnist & 0.9999 & 1.0000 & 0.9999 & 0.9996 & 1.0000 & 0.9998 & 1.0000 & 1.0000 \\
& & tiny\_imagenet\_200 & 0.9703 & 0.9578 & 0.9708 & 0.9089 & 0.9637 & 0.9792 & 0.9609 & 0.9729 \\
& cifar100 & cifar10 & 0.7378 & 0.6807 & 0.7354 & 0.6265 & 0.6951 & 0.7635 & 0.6853 & 0.7046 \\
& & dtd & 0.9680 & 0.9997 & 0.9998 & 0.9879 & 0.9859 & 0.9961 & 0.9998 & 0.9998 \\
& & letters & 0.9711 & 1.0000 & 1.0000 & 0.9934 & 0.9840 & 0.9935 & 1.0000 & 1.0000 \\
& & mnist & 0.9707 & 1.0000 & 1.0000 & 0.9938 & 1.0000 & 0.9997 & 1.0000 & 1.0000 \\
& & tiny\_imagenet\_200 & 0.9696 & 0.9996 & 0.9997 & 0.9926 & 0.9963 & 0.9973 & 0.9997 & 0.9995 \\
& super\_cifar100-ID & super\_cifar100-OOD & 0.7752 & 0.7534 & 0.7701 & 0.6752 & 0.7678 & 0.8062 & 0.7395 & 0.7431 \\
without ood & cifar10 & cifar100 & 0.9251 & 0.8498 & 0.9253 & 0.7873 & 0.9316 & 0.9509 & 0.8937 & 0.8933 \\
& & dtd & 0.9895 & 0.9393 & 0.9892 & 0.9373 & 0.9883 & 0.9986 & 0.9872 & 0.9894 \\
& & letters & 0.9483 & 0.8418 & 0.9351 & 0.7888 & 0.9474 & 0.9742 & 0.9510 & 0.9773 \\
& & mnist & 0.9455 & 0.8615 & 0.9380 & 0.7739 & 0.9442 & 0.9742 & 0.9322 & 0.9811 \\
& & tiny\_imagenet\_200 & 0.9218 & 0.8534 & 0.9217 & 0.7854 & 0.9262 & 0.9611 & 0.8963 & 0.9273 \\
& cifar100 & cifar10 & 0.7017 & 0.6350 & 0.7004 & 0.5825 & 0.6917 & 0.8224 & 0.6273 & 0.6359 \\
& & dtd & 0.9137 & 0.7765 & 0.9103 & 0.7772 & 0.8758 & 0.9523 & 0.8805 & 0.9268 \\
& & letters & 0.6770 & 0.6419 & 0.6963 & 0.5607 & 0.7172 & 0.7204 & 0.7280 & 0.7716 \\
& & mnist & 0.6591 & 0.6213 & 0.6752 & 0.5573 & 0.7039 & 0.6864 & 0.7047 & 0.7605 \\
& & tiny\_imagenet\_200 & 0.8199 & 0.7407 & 0.8206 & 0.7127 & 0.8095 & 0.8601 & 0.7858 & 0.8202 \\
& super\_cifar100-ID & super\_cifar100-OOD & 0.7789 & 0.7430 & 0.7804 & 0.6813 & 0.7679 & 0.8142 & 0.7546 & 0.7413 \\
\bottomrule
\end{tabular}
\end{adjustbox}
\end{table}

\begin{table}[H]
\centering
\begin{threeparttable}
\caption{Significant Differences in the Benchmark Truth Based on the AUROC Metric}
\label{test_verdad_auroc}
\begin{tabular}{lccccccc}
\toprule
 & ebo & fdbd & gen & klm & knn & mds & nnguide \\
\midrule
ebo      & 1 & \textbf{0.0059}*** & \textbf{0.1465} & \textbf{0.0003}*** & 0.7745 & \textbf{0.0001}*** & \textbf{0.4434} \\
fdbd     & \textbf{0.0059}*** & 1 & \textbf{0.0000}*** & \textbf{0.0000}*** & \textbf{0.0074}*** & \textbf{0.0025}*** & \textbf{0.0019}*** \\
gen      & \textbf{0.1465} & \textbf{0.0000}*** & 1 & \textbf{0.0000}*** & \textbf{0.0984}* & \textbf{0.0066}*** & \textbf{0.0917}* \\
klm      & \textbf{0.0003}*** & \textbf{0.0000}*** & \textbf{0.0000}*** & 1 & \textbf{0.0001}*** & \textbf{0.0000}*** & \textbf{0.0000}*** \\
knn      & 0.7745 & \textbf{0.0074}*** & \textbf{0.0984}* & \textbf{0.0001}*** & 1 & \textbf{0.0001}*** & \textbf{0.5661} \\
mds      & \textbf{0.0001}*** & \textbf{0.0025}*** & \textbf{0.0066}*** & \textbf{0.0000}*** & \textbf{0.0001}*** & 1 & \textbf{0.0301}** \\
nnguide  & \textbf{0.4434} & \textbf{0.0019}*** & \textbf{0.0917}* & \textbf{0.0000}*** & \textbf{0.5661} & \textbf{0.0301}** & 1 \\
\bottomrule
\end{tabular}
\begin{tablenotes}
\footnotesize
\item \textbf{Note:} Significance levels are indicated as follows: * $p < 0.1$, ** $p < 0.05$, *** $p < 0.01$.
\end{tablenotes}
\end{threeparttable}
\end{table}

\begin{table}[H]
\centering
\begin{threeparttable}
\caption{Hit Rate for the Metric auroc\_pval$\leq$0.1}
\label{aciertos_auroc0.1}
\begin{tabular}{l|cccccccc}
\textbf{} & \textbf{ebo} & \textbf{fdbd} & \textbf{gen} & \textbf{klm} & \textbf{knn} & \textbf{mds} & \textbf{nnguide} & \textbf{relation} \\
\hline
\textbf{ebo}      & 0 & \textbf{10} & -2 & \textbf{10} & 0 & \textbf{10} & 0 & \textbf{6} \\
\textbf{fdbd}     & \textbf{10} & 0 & \textbf{10} & \textbf{10} & \textbf{10} & \textbf{10} & \textbf{10} & -5 \\
\textbf{gen}      & -2 & \textbf{10} & 0 & \textbf{10} & \textbf{1} & \textbf{9} & \textbf{3} & \textbf{9} \\
\textbf{klm}      & \textbf{10} & \textbf{10} & \textbf{10} & 0 & \textbf{10} & \textbf{10} & \textbf{10} & \textbf{10} \\
\textbf{knn}      & 0 & \textbf{10} & \textbf{1} & \textbf{10} & 0 & \textbf{10} & -1 & \textbf{8} \\
\textbf{mds}      & \textbf{10} & \textbf{10} & \textbf{9} & \textbf{10} & \textbf{10} & 0 & \textbf{9} & \textbf{10} \\
\textbf{nnguide}  & 0 & \textbf{10} & \textbf{3} & \textbf{10} & -1 & \textbf{9} & 0 & \textbf{10} \\
\textbf{relation} & \textbf{6} & -5 & \textbf{9} & \textbf{10} & \textbf{8} & \textbf{10} & \textbf{10} & 0 \\

\end{tabular}
\begin{tablenotes}
\footnotesize
\item \textbf{Note:} Bold values indicate method pairs with significant differences also found in the benchmark truth. Negative signs denote significant differences found in DCV-ROOD but not in the benchmark truth.
\end{tablenotes}
\end{threeparttable}
\end{table}

\begin{table}[H]
\centering
\begin{threeparttable}
\caption{Hit Rate for the Metric auroc\_pval$\leq$0.05}
\label{aciertos_auroc0.5}
\begin{tabular}{l|cccccccc}
\textbf{} & \textbf{ebo} & \textbf{fdbd} & \textbf{gen} & \textbf{klm} & \textbf{knn} & \textbf{mds} & \textbf{nnguide} & \textbf{relation} \\
\hline
\textbf{ebo}      & - & \textbf{10} & 0 & \textbf{10} & 0 & \textbf{10} & 0 & \textbf{6} \\
\textbf{fdbd}     & \textbf{10} & - & \textbf{10} & \textbf{10} & \textbf{9} & \textbf{10} & \textbf{10} & -1 \\
\textbf{gen}      & 0 & \textbf{10} & - & \textbf{10} & -1 & \textbf{8} & -3 & \textbf{8} \\
\textbf{klm}      & \textbf{10} & \textbf{10} & \textbf{10} & - & \textbf{10} & \textbf{10} & \textbf{10} & \textbf{10} \\
\textbf{knn}      & 0 & \textbf{9} & -1 & \textbf{10} & - & \textbf{9} & -1 & \textbf{7} \\
\textbf{mds}      & \textbf{10} & \textbf{10} & \textbf{8} & \textbf{10} & \textbf{9} & - & \textbf{8} & \textbf{9} \\
\textbf{nnguide}  & 0 & \textbf{10} & -3 & \textbf{10} & -1 & \textbf{8} & - & \textbf{10} \\
\textbf{relation} & \textbf{6} & -1 & \textbf{8} & \textbf{10} & \textbf{7} & \textbf{9} & \textbf{10} & - \\
\end{tabular}
\begin{tablenotes}
\footnotesize
\item \textbf{Note:} Bold values indicate method pairs with significant differences also found in the benchmark truth. Negative signs denote significant differences found in DCV-ROOD but not in the benchmark truth.
\end{tablenotes}
\end{threeparttable}
\end{table}

\subsection{AUPR}
\label{aupr}

\begin{table}[H]
\centering
\caption{Benchmark Truth (AUPR on OOD Detection)}
\label{aupr_benchmark}
\begin{adjustbox}{width=\textwidth}
\begin{tabular}{lllcccccccc}
\toprule
versions & id\_database & ood\_database & ebo & fdbd & gen & klm & knn & mds & nnguide & relation \\
\midrule
with ood & cifar10 & cifar100 & 0.9322 & 0.9255 & 0.9366 & 0.7676 & 0.9052 & 0.9142 & 0.9182 & 0.9301 \\
         &         & dtd & 0.9999 & 0.9999 & 1.0000 & 0.9995 & 0.9995 & 0.9998 & 1.0000 & 0.9999 \\
         &         & letters & 0.9993 & 0.9993 & 0.9993 & 0.9970 & 0.9987 & 0.9989 & 0.9994 & 0.9994 \\
         &         & mnist & 1.0000 & 1.0000 & 1.0000 & 0.9997 & 1.0000 & 0.9999 & 1.0000 & 0.9999 \\
         &         & tiny\_imagenet\_200 & 0.9345 & 0.9059 & 0.9364 & 0.7888 & 0.9336 & 0.9576 & 0.9069 & 0.9204 \\
         & cifar100 & cifar10 & 0.7324 & 0.6804 & 0.7319 & 0.5966 & 0.6808 & 0.7648 & 0.6726 & 0.6935 \\
         &          & dtd & 0.9980 & 1.0000 & 1.0000 & 0.9992 & 0.9979 & 0.9997 & 1.0000 & 1.0000 \\
         &          & letters & 0.9786 & 1.0000 & 1.0000 & 0.9951 & 0.9863 & 0.9947 & 1.0000 & 1.0000 \\
         &          & mnist & 0.9840 & 1.0000 & 1.0000 & 0.9963 & 1.0000 & 0.9997 & 1.0000 & 1.0000 \\
         &          & tiny\_imagenet\_200 & 0.9767 & 0.9986 & 0.9988 & 0.9895 & 0.9960 & 0.9966 & 0.9990 & 0.9989 \\
         & super\_cifar100-ID & super\_cifar100-OOD & 0.9292 & 0.9215 & 0.9276 & 0.8648 & 0.9291 & 0.9407 & 0.9164 & 0.8917 \\
without ood & cifar10 & cifar100 & 0.9277 & 0.8279 & 0.9297 & 0.6791 & 0.9392 & 0.9540 & 0.8805 & 0.9046 \\
         &         & dtd & 0.9992 & 0.9946 & 0.9992 & 0.9905 & 0.9991 & 0.9999 & 0.9990 & 0.9990 \\
         &         & letters & 0.9075 & 0.6947 & 0.8786 & 0.4959 & 0.9068 & 0.9483 & 0.9119 & 0.9329 \\
         &         & mnist & 0.9517 & 0.8562 & 0.9437 & 0.6611 & 0.9520 & 0.9761 & 0.9360 & 0.9574 \\
         &         & tiny\_imagenet\_200 & 0.8121 & 0.6533 & 0.8171 & 0.4430 & 0.8352 & 0.9029 & 0.7314 & 0.7581 \\
         & cifar100 & cifar10 & 0.6689 & 0.6485 & 0.6761 & 0.5118 & 0.6865 & 0.8299 & 0.5909 & 0.6047 \\
         &          & dtd & 0.9924 & 0.9753 & 0.9921 & 0.9685 & 0.9890 & 0.9963 & 0.9886 & 0.9812 \\
         &          & letters & 0.4539 & 0.5192 & 0.4943 & 0.3364 & 0.5318 & 0.5183 & 0.5984 & 0.7231 \\
         &          & mnist & 0.6151 & 0.6535 & 0.6443 & 0.5271 & 0.6776 & 0.6460 & 0.7137 & 0.8067 \\
         &          & tiny\_imagenet\_200 & 0.6271 & 0.4862 & 0.6223 & 0.4213 & 0.6473 & 0.7263 & 0.5550 & 0.4811 \\
         & super\_cifar100-ID & super\_cifar100-OOD & 0.9287 & 0.9150 & 0.9294 & 0.8656 & 0.9255 & 0.9407 & 0.9172 & 0.8764 \\
\bottomrule
\end{tabular}
\end{adjustbox}
\end{table}

\begin{table}[H]
\centering
\begin{threeparttable}
\caption{Significant Differences in the Benchmark Truth Based for the AUPR Metric}
\label{test_verdad_aupr}
\begin{tabular}{lcccccccc}
\toprule
 & ebo & fdbd & gen & klm & knn & mds & nnguide & relation \\
\midrule
ebo      & 1 & \textbf{0.0794}* & \textbf{0.0738}* & \textbf{0.0004}*** & 0.3880 & \textbf{0.0001}*** & \textbf{0.4060} & \textbf{0.0917}* \\
fdbd     & \textbf{0.0794}* & 1 & \textbf{0.0013}*** & \textbf{0.0000}*** & \textbf{0.0127}*** & \textbf{0.0229}** & \textbf{0.0141}*** & \textbf{0.5879} \\
gen      & \textbf{0.0738}* & \textbf{0.0013}*** & 1 & \textbf{0.0000}*** & \textbf{0.7502} & \textbf{0.0083}*** & 0.1659 & \textbf{0.0011}*** \\
klm      & \textbf{0.0004}*** & \textbf{0.0000}*** & \textbf{0.0000}*** & 1 & \textbf{0.0000}*** & \textbf{0.0000}*** & \textbf{0.0000}*** & \textbf{0.0000}*** \\
knn      & 0.3880 & \textbf{0.0127}*** & 0.7502 & \textbf{0.0000}*** & 1 & \textbf{0.0015}*** & \textbf{0.4245} & \textbf{0.0462}** \\
mds      & \textbf{0.0001}*** & \textbf{0.0229}** & \textbf{0.0083}*** & \textbf{0.0000}*** & \textbf{0.0015}*** & 1 & \textbf{0.0854}* & \textbf{0.0329}** \\
nnguide  & 0.4060 & \textbf{0.0141}*** & 0.1659 & \textbf{0.0000}*** & 0.4245 & \textbf{0.0854}* & 1 & \textbf{0.0011}*** \\
relation & \textbf{0.0917}* & 0.5879 & \textbf{0.0011}*** & \textbf{0.0000}*** & \textbf{0.0462}** & \textbf{0.0329}** & \textbf{0.0011}*** & 1 \\
\bottomrule
\end{tabular}
\begin{tablenotes}
\footnotesize
\item \textbf{Note:} Significance levels are indicated as follows: * $p < 0.1$, ** $p < 0.05$, *** $p < 0.01$.
\end{tablenotes}
\end{threeparttable}
\end{table}

\begin{table}[H]
\centering
\begin{threeparttable}
\caption{Hit Rate for the Metric aupr\_pval$\leq$0.1}
\label{aciertos_aupr}
\begin{tabular}{l|cccccccc}
\textbf{} & \textbf{ebo} & \textbf{fdbd} & \textbf{gen} & \textbf{klm} & \textbf{knn} & \textbf{mds} & \textbf{nnguide} & \textbf{relation} \\
\hline
\textbf{ebo}      & 0 & \textbf{9} & 0 & \textbf{10} & 0 & \textbf{10} & 0 & \textbf{4} \\
\textbf{fdbd}     & \textbf{9} & 0 & \textbf{10} & \textbf{10} & \textbf{10} & \textbf{10} & \textbf{10} & 0 \\
\textbf{gen}      & 0 & \textbf{10} & 0 & \textbf{10} & 0 & \textbf{9} & -3 & \textbf{8} \\
\textbf{klm}      & \textbf{10} & \textbf{10} & \textbf{10} & 0 & \textbf{10} & \textbf{10} & \textbf{10} & \textbf{10} \\
\textbf{knn}      & 0 & \textbf{10} & 0 & \textbf{10} & 0 & \textbf{9} & 0 & \textbf{9} \\
\textbf{mds}      & \textbf{10} & \textbf{10} & \textbf{9} & \textbf{10} & \textbf{9} & 0 & \textbf{8} & \textbf{9} \\
\textbf{nnguide}  & 0 & \textbf{10} & -3 & \textbf{10} & 0 & \textbf{8} & 0 & \textbf{10} \\
\textbf{relation} & \textbf{4} & 0 & \textbf{8} & \textbf{10} & \textbf{9} & \textbf{9} & \textbf{10} & 0 \\
\end{tabular}
\begin{tablenotes}
\footnotesize
\item \textbf{Note:} Bold values indicate method pairs with significant differences also found in the benchmark truth. Negative signs denote significant differences found in DCV-ROOD but not in the benchmark truth.
\end{tablenotes}
\end{threeparttable}
\end{table}

\begin{table}[H]
\centering
\begin{threeparttable}
\caption{Hit Rate for the Metric aupr\_pval$\leq$0.05}
\label{aciertos_aupr_005}
\begin{tabular}{l|cccccccc}
\textbf{} & \textbf{ebo} & \textbf{fdbd} & \textbf{gen} & \textbf{klm} & \textbf{knn} & \textbf{mds} & \textbf{nnguide} & \textbf{relation} \\
\hline
\textbf{ebo}      & 0 & -6 & 0 & \textbf{10} & 0 & \textbf{10} & 0 & -3 \\
\textbf{fdbd}     & -6 & 0 & \textbf{10} & \textbf{10} & \textbf{10} & \textbf{10} & \textbf{10} & 0 \\
\textbf{gen}      & 0 & \textbf{10} & 0 & \textbf{10} & 0 & \textbf{8} & -2 & \textbf{5} \\
\textbf{klm}      & \textbf{10} & \textbf{10} & \textbf{10} & 0 & \textbf{10} & \textbf{10} & \textbf{10} & \textbf{10} \\
\textbf{knn}      & 0 & \textbf{10} & 0 & \textbf{10} & 0 & \textbf{7} & 0 & \textbf{9} \\
\textbf{mds}      & \textbf{10} & \textbf{10} & \textbf{8} & \textbf{10} & \textbf{7} & 0 & -6 & \textbf{8} \\
\textbf{nnguide}  & 0 & \textbf{10} & -2 & \textbf{10} & 0 & \textbf{6} & 0 & \textbf{10} \\
\textbf{relation} & -3 & 0 & \textbf{5} & \textbf{10} & \textbf{9} & \textbf{8} & \textbf{10} & 0 \\
\end{tabular}
\begin{tablenotes}
\footnotesize
\item \textbf{Note:} Bold values indicate method pairs with significant differences also found in the benchmark truth. Negative signs denote significant differences found in DCV-ROOD but not in the benchmark truth.
\end{tablenotes}
\end{threeparttable}
\end{table}

\subsection{F1-Score}
\label{f1}

\begin{table}[H]
\centering
\caption{Benchmark Truth (F1-Score on OOD Detection)}
\label{f1_benchmark}
\begin{adjustbox}{width=\textwidth}
\begin{tabular}{lllcccccccc}
\toprule
versions & id\_database & ood\_database & ebo & fdbd & gen & klm & knn & mds & nnguide & relation \\
\midrule
with ood & cifar10 & cifar100 & 0.8569 & 0.8513 & 0.8567 & 0.8440 & 0.8283 & 0.8176 & 0.8484 & 0.8419 \\
& & dtd & 0.9993 & 0.9996 & 0.9996 & 0.9987 & 0.9982 & 0.9976 & 0.9996 & 0.9997 \\
& & letters & 0.9986 & 0.9985 & 0.9985 & 0.9980 & 0.9984 & 0.9982 & 0.9987 & 0.9992 \\
& & mnist & 0.9997 & 0.9999 & 0.9998 & 0.9990 & 0.9997 & 0.9997 & 0.9998 & 0.9999 \\
& & tiny\_imagenet\_200 & 0.9140 & 0.8985 & 0.9137 & 0.8705 & 0.9094 & 0.9230 & 0.9002 & 0.9152 \\
& cifar100 & cifar10 & 0.7188 & 0.6973 & 0.7201 & 0.6954 & 0.7006 & 0.7235 & 0.7094 & 0.7122 \\
& & dtd & 0.9839 & 0.9998 & 0.9998 & 0.9944 & 0.9991 & 0.9987 & 0.9998 & 0.9997 \\
& & letters & 0.9853 & 0.9999 & 0.9999 & 0.9962 & 0.9885 & 0.9950 & 0.9999 & 0.9999 \\
& & mnist & 0.9844 & 0.9997 & 0.9995 & 0.9946 & 0.9998 & 0.9980 & 0.9999 & 0.9997 \\
& & tiny\_imagenet\_200 & 0.9809 & 0.9959 & 0.9955 & 0.9894 & 0.9938 & 0.9917 & 0.9963 & 0.9928 \\
& super\_cifar100-ID & super\_cifar100-OOD & 0.8922 & 0.8922 & 0.8923 & 0.8894 & 0.8901 & 0.8953 & 0.8919 & 0.8929 \\
without ood & cifar10 & cifar100 & 0.8506 & 0.7952 & 0.8481 & 0.8077 & 0.8577 & 0.8785 & 0.8335 & 0.8235 \\
& & dtd & 0.9916 & 0.9764 & 0.9916 & 0.9757 & 0.9895 & 0.9973 & 0.9905 & 0.9938 \\
& & letters & 0.8175 & 0.6672 & 0.7865 & 0.6542 & 0.8157 & 0.8721 & 0.8352 & 0.8867 \\
& & mnist & 0.8791 & 0.7980 & 0.8682 & 0.8000 & 0.8815 & 0.9231 & 0.8648 & 0.9387 \\
& & tiny\_imagenet\_200 & 0.7585 & 0.6784 & 0.7559 & 0.6647 & 0.7696 & 0.8286 & 0.7284 & 0.7658 \\
& cifar100 & cifar10 & 0.7187 & 0.6847 & 0.7168 & 0.6973 & 0.7001 & 0.7541 & 0.7056 & 0.6927 \\
& & dtd & 0.9771 & 0.9694 & 0.9763 & 0.9695 & 0.9691 & 0.9817 & 0.9744 & 0.9802 \\
& & letters & 0.5205 & 0.4827 & 0.5264 & 0.4702 & 0.5262 & 0.5302 & 0.5493 & 0.5835 \\
& & mnist & 0.6788 & 0.6482 & 0.6816 & 0.6585 & 0.6882 & 0.6791 & 0.6827 & 0.7262 \\
& & tiny\_imagenet\_200 & 0.6212 & 0.5387 & 0.6255 & 0.5411 & 0.6103 & 0.6676 & 0.5864 & 0.6147 \\
& super\_cifar100-ID & super\_cifar100-OOD & 0.8947 & 0.8912 & 0.8946 & 0.8907 & 0.8916 & 0.9004 & 0.8930 & 0.8941 \\
\bottomrule
\end{tabular}
\end{adjustbox}
\end{table}

\begin{table}[H]
\centering
\begin{threeparttable}
\caption{Significant Differences in the Benchmark Truth Based for the F1-Score Metric}
\label{test_verdad_f1}
\begin{tabular}{lcccccccc}
\toprule
 & ebo & fdbd & gen & klm & knn & mds & nnguide & relation \\
\midrule
ebo      & 1 & \textbf{0.0103}*** & 0.4628 & \textbf{0.0017}*** & 0.9240 & \textbf{0.0005}*** & 0.7502 & \textbf{0.0794}* \\
fdbd     & \textbf{0.0103}*** & 1 & \textbf{0.0003}*** & \textbf{0.0684}* & \textbf{0.0501}** & \textbf{0.0066}*** & \textbf{0.0002}*** & \textbf{0.0229}** \\
gen      & 0.4628 & \textbf{0.0003}*** & 1 & \textbf{0.0000}*** & 0.1870 & \textbf{0.0275}** & \textbf{0.1289} & \textbf{0.0002}*** \\
klm      & \textbf{0.0017}*** & \textbf{0.0684}* & \textbf{0.0000}*** & 1 & \textbf{0.0011}*** & \textbf{0.0002}*** & \textbf{0.0000}*** & \textbf{0.0000}*** \\
knn      & 0.9240 & \textbf{0.0501}** & 0.1870 & \textbf{0.0011}*** & 1 & \textbf{0.0074}*** & 0.5028 & 0.6327 \\
mds      & \textbf{0.0005}*** & \textbf{0.0066}*** & \textbf{0.0275}** & \textbf{0.0002}*** & \textbf{0.0074}*** & 1 & \textbf{0.0634}* & \textbf{0.0301}** \\
nnguide  & 0.7502 & \textbf{0.0002}*** & 0.1289 & \textbf{0.0000}*** & 0.5028 & \textbf{0.0634}* & 1 & \textbf{0.0008}*** \\
relation & \textbf{0.0794}* & \textbf{0.0229}** & \textbf{0.0002}*** & \textbf{0.0000}*** & 0.6327 & \textbf{0.0301}** & \textbf{0.0008}*** & 1 \\
\bottomrule
\end{tabular}
\begin{tablenotes}
\footnotesize
\item \textbf{Note:} Significance levels are indicated as follows: * $p < 0.1$, ** $p < 0.05$, *** $p < 0.01$.
\end{tablenotes}
\end{threeparttable}
\end{table}

\begin{table}[H]
\centering
\begin{threeparttable}
\caption{Hit Rate for the Metric F1\_score\_pval$\leq$0.1}
\label{aciertos_f1_01}
\begin{tabular}{l|cccccccc}
\textbf{} & \textbf{ebo} & \textbf{fdbd} & \textbf{gen} & \textbf{klm} & \textbf{knn} & \textbf{mds} & \textbf{nnguide} & \textbf{relation} \\
\hline
\textbf{ebo}      & 0 & \textbf{10} & 0 & \textbf{10} & 0 & \textbf{9} & 0 & \textbf{3} \\
\textbf{fdbd}     & \textbf{10} & 0 & \textbf{10} & 2 & \textbf{10} & \textbf{10} & \textbf{10} & \textbf{10} \\
\textbf{gen}      & 0 & \textbf{10} & 0 & \textbf{10} & -1 & \textbf{9} & -2 & -6 \\
\textbf{klm}      & \textbf{10} & 2 & \textbf{10} & 0 & \textbf{10} & \textbf{10} & \textbf{10} & \textbf{10} \\
\textbf{knn}      & 0 & \textbf{10} & -1 & \textbf{10} & 0 & \textbf{10} & 0 & -3 \\
\textbf{mds}      & \textbf{9} & \textbf{10} & \textbf{9} & \textbf{10} & \textbf{10} & 0 & \textbf{9} & \textbf{9} \\
\textbf{nnguide}  & 0 & \textbf{10} & -2 & \textbf{10} & 0 & \textbf{9} & 0 & \textbf{8} \\
\textbf{relation} & \textbf{3} & \textbf{10} & -6 & \textbf{10} & -3 & \textbf{9} & \textbf{8} & 0 \\
\end{tabular}
\begin{tablenotes}
\footnotesize
\item \textbf{Note:} Bold values indicate method pairs with significant differences also found in the benchmark truth. Negative signs denote significant differences found in DCV-ROOD but not in the benchmark truth.
\end{tablenotes}
\end{threeparttable}
\end{table}

\begin{table}[H]
\centering
\begin{threeparttable}
\caption{Hit Rate for the Metric F1\_score\_pval$\leq$0.05}
\label{aciertos_f1_005}
\begin{tabular}{l|cccccccc}
\textbf{} & \textbf{ebo} & \textbf{fdbd} & \textbf{gen} & \textbf{klm} & \textbf{knn} & \textbf{mds} & \textbf{nnguide} & \textbf{relation} \\
\hline
\textbf{ebo}      & 0 & \textbf{10} & 0 & \textbf{10} & 0 & \textbf{9} & 0 & -2 \\
\textbf{fdbd}     & \textbf{10} & 0 & \textbf{10} & -1 & -10 & \textbf{10} & \textbf{10} & \textbf{10} \\
\textbf{gen}      & 0 & \textbf{10} & 0 & \textbf{10} & 0 & \textbf{8} & -1 & \textbf{5} \\
\textbf{klm}      & \textbf{10} & -1 & \textbf{10} & 0 & \textbf{10} & \textbf{10} & \textbf{10} & \textbf{10} \\
\textbf{knn}      & 0 & -10 & 0 & \textbf{10} & 0 & \textbf{10} & 0 & -2 \\
\textbf{mds}      & \textbf{9} & \textbf{10} & \textbf{8} & \textbf{10} & \textbf{10} & 0 & -7 & \textbf{8} \\
\textbf{nnguide}  & 0 & \textbf{10} & -1 & \textbf{10} & 0 & -7 & 0 & \textbf{5} \\
\textbf{relation} & -2 & \textbf{10} & \textbf{5} & \textbf{10} & -2 & \textbf{8} & \textbf{5} & 0 \\
\end{tabular}
\begin{tablenotes}
\footnotesize
\item \textbf{Note:} Bold values indicate method pairs with significant differences also found in the benchmark truth. Negative signs denote significant differences found in DCV-ROOD but not in the benchmark truth.
\end{tablenotes}
\end{threeparttable}
\end{table}

\subsection{Accuracy}
\label{acc}

\begin{table}[H]
\centering
\caption{Benchmark Truth (Accuracy\_90 on OOD Detection)}
\label{accuracy_90}

\begin{adjustbox}{width=\textwidth}
\begin{tabular}{lllcccccccc}
\toprule
versions & id\_database & ood\_database & ebo & fdbd & gen & klm & knn & mds & nnguide & relation \\
\midrule
with ood         & cifar10          & cifar100           & 0.8407 & 0.8353 & 0.8423 & 0.8176 & 0.6725 & 0.7581 & 0.8327 & 0.8373 \\
                 &                  & dtd                & 0.9061 & 0.9061 & 0.9060 & 0.9057 & 0.9040 & 0.9056 & 0.9061 & 0.9057 \\
                 &                  & letters            & 0.9730 & 0.9728 & 0.9729 & 0.9728 & 0.9698 & 0.9710 & 0.9730 & 0.9730 \\
                 &                  & mnist              & 0.9537 & 0.9538 & 0.9538 & 0.9537 & 0.9537 & 0.9539 & 0.9537 & 0.9538 \\
                 &                  & tiny\_imagenet\_200 & 0.9124 & 0.8852 & 0.9092 & 0.8811 & 0.8938 & 0.9296 & 0.9030 & 0.8858 \\
                 & cifar100         & cifar10            & 0.6344 & 0.5998 & 0.6363 & 0.5809 & 0.6058 & 0.6455 & 0.6146 & 0.6152 \\
                 &                  & dtd                & 0.9068 & 0.9065 & 0.9066 & 0.8961 & 0.9050 & 0.9058 & 0.9065 & 0.9062 \\
                 &                  & letters            & 0.9738 & 0.9739 & 0.9737 & 0.9711 & 0.9590 & 0.9591 & 0.9738 & 0.9737 \\
                 &                  & mnist              & 0.9541 & 0.9541 & 0.9541 & 0.9489 & 0.9538 & 0.9541 & 0.9542 & 0.9539 \\
                 &                  & tiny\_imagenet\_200 & 0.9717 & 0.9720 & 0.9719 & 0.9686 & 0.9646 & 0.9644 & 0.9720 & 0.9716 \\
                 & super\_cifar100-ID  & super\_cifar100-OOD    & 0.7938 & 0.7843 & 0.7896 & 0.7781 & 0.7864 & 0.8079 & 0.7770 & 0.7901 \\
without ood         & cifar10          & cifar100           & 0.8345 & 0.7480 & 0.8313 & 0.7771 & 0.8420 & 0.8735 & 0.8154 & 0.7833 \\
                 &                  & dtd                & 0.9045 & 0.8963 & 0.9046 & 0.9013 & 0.9047 & 0.9060 & 0.9046 & 0.9021 \\
                 &                  & letters            & 0.8455 & 0.6690 & 0.8195 & 0.7230 & 0.8486 & 0.9176 & 0.8698 & 0.7790 \\
                 &                  & mnist              & 0.8552 & 0.7368 & 0.8428 & 0.7755 & 0.8519 & 0.9188 & 0.8409 & 0.7992 \\
                 &                  & tiny\_imagenet\_200 & 0.8171 & 0.6966 & 0.8118 & 0.7366 & 0.8226 & 0.8934 & 0.7961 & 0.7455 \\
                 & cifar100         & cifar10            & 0.6444 & 0.5742 & 0.6416 & 0.6099 & 0.6108 & 0.6960 & 0.6206 & 0.6074 \\
                 &                  & dtd                & 0.8911 & 0.8754 & 0.8908 & 0.8728 & 0.8850 & 0.8975 & 0.8877 & 0.8829 \\
                 &                  & letters            & 0.5166 & 0.4280 & 0.5257 & 0.4460 & 0.5201 & 0.5204 & 0.5191 & 0.4697 \\
                 &                  & mnist              & 0.5827 & 0.5242 & 0.5904 & 0.5549 & 0.6062 & 0.5819 & 0.5900 & 0.5471 \\
                 &                  & tiny\_imagenet\_200 & 0.6201 & 0.5164 & 0.6236 & 0.5420 & 0.5844 & 0.6741 & 0.5826 & 0.5395 \\
                 & super\_cifar100-ID  & super\_cifar100-OOD    & 0.8021 & 0.7859 & 0.8019 & 0.7852 & 0.7954 & 0.8167 & 0.7950 & 0.7915 \\
\bottomrule
\end{tabular}
\end{adjustbox}
\end{table}

\begin{table}[H]
\centering
\begin{threeparttable}
\caption{Significant Differences in the Benchmark Truth Based for the Accuracy\_90 Metric}
\label{test_verdad_acc}
\begin{tabular}{lcccccccc}
\toprule
 & ebo & fdbd & gen & klm & knn & mds & nnguide & relation \\
\midrule
ebo      & 1 & \textbf{0.0001}*** & 0.3705 & \textbf{0.0000}*** & \textbf{0.0301}** & \textbf{0.0462}** & \textbf{0.0537}* & \textbf{0.0000}*** \\
fdbd     & \textbf{0.0001}*** & 1 & \textbf{0.0002}*** & 0.3535 & \textbf{0.0251}** & \textbf{0.0042}*** & \textbf{0.0009}*** & \textbf{0.0019}*** \\
gen      & 0.3705 & \textbf{0.0002}*** & 1 & \textbf{0.0000}*** & 0.1055* & \textbf{0.0794}* & \textbf{0.0115}** & \textbf{0.0000}*** \\
klm      & \textbf{0.0000}*** & 0.3535 & \textbf{0.0000}*** & 1 & \textbf{0.0074}*** & \textbf{0.0009}*** & \textbf{0.0000}*** & \textbf{0.0005}*** \\
knn      & \textbf{0.0301}** & \textbf{0.0251}** & 0.1055* & \textbf{0.0074}*** & 1 & \textbf{0.0001}*** & 0.5446 & 0.1465 \\
mds      & \textbf{0.0462}** & \textbf{0.0042}*** & \textbf{0.0794}* & \textbf{0.0009}*** & \textbf{0.0001}*** & 1 & \textbf{0.0329}** & \textbf{0.0047}*** \\
nnguide  & \textbf{0.0537}* & \textbf{0.0009}*** & \textbf{0.0115}** & \textbf{0.0000}*** & 0.5446 & \textbf{0.0329}** & 1 & \textbf{0.0017}*** \\
relation & \textbf{0.0000}*** & \textbf{0.0019}*** & \textbf{0.0000}*** & \textbf{0.0005}*** & 0.1465 & \textbf{0.0047}*** & \textbf{0.0017}*** & 1 \\
\bottomrule
\end{tabular}
\begin{tablenotes}
\footnotesize
\item \textbf{Note:} Significance levels are indicated as follows: * $p < 0.1$, ** $p < 0.05$, *** $p < 0.01$.
\end{tablenotes}
\end{threeparttable}
\end{table}

\begin{table}[H]
\centering
\begin{threeparttable}
\caption{Hit Rate for the Metric acc\_90\_pval$\leq$0.1}
\label{aciertos_acc90_01}
\begin{tabular}{l|cccccccc}
\textbf{} & \textbf{ebo} & \textbf{fdbd} & \textbf{gen} & \textbf{klm} & \textbf{knn} & \textbf{mds} & \textbf{nnguide} & \textbf{relation} \\
\hline
\textbf{ebo}      & 0 & \textbf{10} & 0 & \textbf{10} & \textbf{4} & \textbf{8} & \textbf{2} & \textbf{10} \\
\textbf{fdbd}     & \textbf{10} & 0 & \textbf{10} & -2 & \textbf{9} & \textbf{10} & \textbf{10} & \textbf{10} \\
\textbf{gen}      & 0 & \textbf{10} & 0 & \textbf{10} & -1 & \textbf{8} & \textbf{3} & \textbf{10} \\
\textbf{klm}      & \textbf{10} & -2 & \textbf{10} & 0 & \textbf{9} & \textbf{10} & \textbf{10} & \textbf{10} \\
\textbf{knn}      & \textbf{4} & \textbf{9} & -1 & \textbf{9} & 0 & \textbf{10} & 0 & -5 \\
\textbf{mds}      & \textbf{8} & \textbf{10} & \textbf{8} & \textbf{10} & \textbf{10} & 0 & \textbf{7} & \textbf{9} \\
\textbf{nnguide}  & \textbf{2} & \textbf{10} & \textbf{3} & \textbf{10} & 0 & \textbf{7} & 0 & \textbf{9} \\
\textbf{relation} & \textbf{10} & \textbf{10} & \textbf{10} & \textbf{10} & -5 & \textbf{9} & \textbf{9} & 0 \\
\end{tabular}
\begin{tablenotes}
\footnotesize
\item \textbf{Note:} Bold values indicate method pairs with significant differences also found in the benchmark truth. Negative signs denote significant differences found in DCV-ROOD but not in the benchmark truth.
\end{tablenotes}
\end{threeparttable}
\end{table}

\begin{table}[H]
\centering
\begin{threeparttable}
\caption{Hit Rate for the Metric acc\_90\_pval$\leq$0.05}
\label{aciertos_acc90_005}
\begin{tabular}{l|cccccccc}
\textbf{} & \textbf{ebo} & \textbf{fdbd} & \textbf{gen} & \textbf{klm} & \textbf{knn} & \textbf{mds} & \textbf{nnguide} & \textbf{relation} \\
\hline
\textbf{ebo}      & 0 & \textbf{10} & 0 & \textbf{10} & \textbf{2} & \textbf{7} & -1 & \textbf{10} \\
\textbf{fdbd}     & \textbf{10} & 0 & \textbf{10} & -2 & \textbf{5} & \textbf{10} & \textbf{10} & \textbf{9} \\
\textbf{gen}      & 0 & \textbf{10} & 0 & \textbf{10} & -1 & -4 & \textbf{1} & \textbf{10} \\
\textbf{klm}      & \textbf{10} & -2 & \textbf{10} & 0 & \textbf{1} & \textbf{10} & \textbf{10} & \textbf{10} \\
\textbf{knn}      & \textbf{2} & \textbf{5} & -1 & \textbf{1} & 0 & \textbf{9} & 0 & -1 \\
\textbf{mds}      & \textbf{7} & \textbf{10} & -4 & \textbf{10} & \textbf{9} & 0 & \textbf{6} & \textbf{9} \\
\textbf{nnguide}  & -1 & \textbf{10} & \textbf{1} & \textbf{10} & 0 & \textbf{6} & 0 & \textbf{9} \\
\textbf{relation} & \textbf{10} & \textbf{9} & \textbf{10} & \textbf{10} & -1 & \textbf{9} & \textbf{9} & 0 \\
\end{tabular}
\begin{tablenotes}
\footnotesize
\item \textbf{Note:} Bold values indicate method pairs with significant differences also found in the benchmark truth. Negative signs denote significant differences found in DCV-ROOD but not in the benchmark truth.
\end{tablenotes}
\end{threeparttable}
\end{table}

\subsection{TPR5}
\label{tpr}

\begin{table}[H]
\centering
\begin{threeparttable}
\caption{Hit Rate for the Metric TPR@FPR\_pval$\leq$0.1}
\label{aciertos_tprfpr_01}
\begin{tabular}{l|cccccccc}
\textbf{} & \textbf{ebo} & \textbf{fdbd} & \textbf{gen} & \textbf{klm} & \textbf{knn} & \textbf{mds} & \textbf{nnguide} & \textbf{relation} \\
\hline
\textbf{ebo}      & 0 & \textbf{10} & 0 & \textbf{10} & -1 & \textbf{10} & -2 & \textbf{7} \\
\textbf{fdbd}     & \textbf{10} & 0 & \textbf{10} & \textbf{10} & \textbf{10} & \textbf{10} & \textbf{10} & 0 \\
\textbf{gen}      & 0 & \textbf{10} & 0 & \textbf{10} & 0 & \textbf{10} & -3 & \textbf{10} \\
\textbf{klm}      & \textbf{10} & \textbf{10} & \textbf{10} & 0 & \textbf{10} & \textbf{10} & \textbf{10} & \textbf{10} \\
\textbf{knn}      & -1 & \textbf{10} & 0 & \textbf{10} & 0 & \textbf{10} & -1 & \textbf{10} \\
\textbf{mds}      & \textbf{10} & \textbf{10} & \textbf{10} & \textbf{10} & \textbf{10} & 0 & \textbf{10} & \textbf{10} \\
\textbf{nnguide}  & -2 & \textbf{10} & -3 & \textbf{10} & -1 & \textbf{10} & 0 & \textbf{10} \\
\textbf{relation} & \textbf{7} & 0 & \textbf{10} & \textbf{10} & \textbf{10} & \textbf{10} & \textbf{10} & 0 \\
\end{tabular}
\begin{tablenotes}
\footnotesize
\item \textbf{Note:} Bold values indicate method pairs with significant differences also found in the benchmark truth. Negative signs denote significant differences found in DCV-ROOD but not in the benchmark truth.
\end{tablenotes}
\end{threeparttable}
\end{table}

\end{document}